\documentclass{article}
\usepackage{graphicx}
\usepackage{amsmath}
\usepackage{amssymb}
\usepackage{mathtools}
\usepackage{amsthm}
\usepackage{algorithm}
\usepackage{algpseudocode}
\usepackage{wrapfig}

    \PassOptionsToPackage{numbers, compress}{natbib}
    \usepackage[preprint]{neurips_2025}

\usepackage[utf8]{inputenc} % allow utf-8 input
\usepackage[T1]{fontenc}    % use 8-bit T1 fonts
\usepackage{hyperref}       % hyperlinks
\usepackage{url}            % simple URL typesetting
\usepackage{booktabs}       % professional-quality tables
\usepackage{amsfonts}       % blackboard math symbols
\usepackage{nicefrac}       % compact symbols for 1/2, etc.
\usepackage{microtype}      % microtypography
\usepackage{xcolor}         % colors
\usepackage{pifont}
\newcommand{\cmark}{\ding{51}}%
\newcommand{\xmark}{\ding{55}}%

\title{IPGO: Indirect Prompt Gradient Optimization for Parameter-Efficient Prompt-level Fine-Tuning on Text-to-Image Models}

\vspace{-18mm}
\author{%
  Jianping Ye$^1$ \quad Michel Wedel$^1$ \quad Kunpeng Zhang$^1$\\
  $^1$University of Maryland, College Park\\
  \texttt{\{jpye00,mwedel,kpzhang\}@umd.edu} \\
}

\begin{document}

\maketitle

\vspace{-6mm}
\begin{abstract}
\vspace{-2mm}

Text-to-Image Diffusion models excel at generating images from text prompts but often exhibit suboptimal alignment with content semantics, aesthetics, and human preferences. To address these limitations, this study proposes a novel parameter-efficient framework, Indirect Prompt Gradient Optimization (IPGO), for prompt-level diffusion model fine-tuning. IPGO enhances prompt embeddings by injecting continuously differentiable embeddings at the beginning and end of the prompt embeddings, leveraging the benefits of low-rank structures combined with the flexibility and nonlinearity offered by rotations. This approach enables gradient-based optimization of injected embeddings under range, orthonormality, and conformity constraints, effectively narrowing the search space, promoting a stable solution, and ensuring alignment between the embeddings of the injected embeddings and the original prompt. In an extension (IPGO+) we add a cross-attention mechanism, which does not involve any additional parameters, on the prompt embedding to enforce dependencies between the original prompt and the inserted embeddings. We conduct extensive evaluations through prompt-wise (IPGO) and prompt-batch (IPGO+) training using three reward models aimed at image aesthetics, image-text alignment, and human preferences across three datasets of varying complexity. The results show that IPGO consistently outperforms state-of-the-art benchmarks, including stable diffusion v1.5 with raw prompts, text-embedding-based methods (TextCraftor), training-based approaches (DRaFT and DDPO), and training-free methods (DPO-Diffusion, Promptist, and ChatGPT-4o). Specifically, IPGO achieves a win-rate exceeding 99\% in prompt-wise learning, and IPGO+ achieves a comparable, but often better performance against current SOTAs (a 75\% win rate) in prompt-batch learning. Furthermore, we illustrate IPGO's generalizability and its capability to significantly enhance image generation quality while requiring minimal training data and limited computational resources.%\footnote{Code is available for reproducibility: \url{https://github.com/Demos750/IPGO}}.
\end{abstract}
\vspace{-4mm}
%\end{abstract}
%\vspace{-4mm}

\section{Introduction}
\label{sec:intro}
Text-to-Image (T2I) Diffusion models have emerged as state-of-the-art pipelines for image generation \citep{liu2024alignment,zhang2023text}, but generated images do not always meet high-quality standards with respect to specific downstream objectives \citep{liu2024alignment}. Further aligning images with their evaluations by humans is often desirable. As a result, users frequently need to experiment with different text prompts to produce images that better align  with aesthetic measures, human preferences, and other criteria \citep{wang2022diffusiondb}. Unfortunately,  many downstream applications lack clear and instructive guidelines for prompt design \cite{liu2022design}, making systematic manual prompt engineering challenging. To mitigate this, several approaches have been proposed \cite{liu2024alignment}, ranging from automatic prompt optimization \cite{hao2024optimizing} to generative model fine-tuning \citep{black2023training,fan2024reinforcement,li2024textcraftor, prabhudesai2023aligning}. However, many of these methods either rely on data-intensive training techniques, such as Supervised Fine-Tuning (SFT) and Reinforcement Learning (RL), or require significant modifications to the generative model itself for each downstream task. Therefore, there remains a keen interest in developing more efficient frameworks for image optimization in alignment tasks.

In this paper, we propose a novel approach to prompt-level optimization, called Indirect Prompt Gradient Optimization (IPGO), for prompt-wise training at inference, and its extension IPGO+ for prompt-batch training, both aimed at improving image qualities. Similar to TextCraftor \citep{li2024textcraftor}, our method relies on fine-tuning of text embeddings through reward guidance. However, instead of fine-tuning the entire text encoder, we \textit{efficiently train only a few injected embeddings} at the beginning and end of the text prompt embeddings. We employ constrained gradient-based optimization on these injected embeddings, keeping both the text encoder and diffusion model frozen during the optimization. As a result, IPGO and IPGO+ are much more parameter-efficient.

To evaluate the effectiveness of our IPGO(+) method, we conduct a comprehensive set of experiments using the Stable Diffusion Model \citep{rombach2022high} on a single L4 GPU with 22.5GB VRAM. We assess performance across three target reward models: image aesthetics \citep{LaionAes}, image-text alignment \citep{radford2021learning}, and human preference scores \citep{wu2023human}. The key contributions of this study are as follows:

(1) We propose IPGO, a novel \textbf{parameter-efficient}, gradient-based approach to prompt optimization in the text embedding space for reward guidance of T2I diffusion models at inference. This approach optimizes carefully constructed new embeddings at both the beginning and end of the prompt text embeddings. IPGO+ extends this by incorporating a cross-attention between the inserted and the original embeddings, without introducing additional parameters.

(2) We show that IPGO(+) is \textbf{effective} in improving image quality for prompt-wise training at inference. Additionally, IPGO+ is able to handle prompt batches for training, achieving better rewards when it is beneficial to share learned inserted embeddings across all prompts within a batch. We demonstrate that IPGO(+) can be applied to a wide range of diffusion models, including SDv1.5, SDXL, and SD3. Moreover, \textbf{convex combinations} of the learned IPGO+ embeddings from different rewards can generate more diverse images.

(3) Our extensive experiments on three different datasets using SDv1.5 and three reward functions show that \textbf{(a)} \textit{for prompt-wise training at inference}, IPGO \textbf{outperforms} seven state-of-the-art gradient-based methods in over \textbf{99\%} of the cases; \textbf{(b)} \textit{for prompt-batch training}, under identical resource allocation (batch size and number of GPUs), IPGO+ outperforms the two best performing state-of-the-art benchmark models (TextCraftor and DRaFT-1) in 75\% of the cases. Overall, IPGO(+) achieves \textbf{competitive} results \textbf{with far fewer parameters}.

\section{Related Work}
\label{sec:related_work}

\paragraph{Text-to-Image Diffusion Probabilistic Models}
Foundational work in T2I generation using diffusion models includes score-based generative models \cite{song2019generative} and diffusion-probabilistic models \cite{sohl2015deep}, and the landmark denoising diffusion probabilistic model \citep[DDPM][]{ho2020denoising}. Subsequent models, such as GLIDE \cite{nichol2021glide} and Imagen \cite{saharia2022photorealistic} operate the diffusion process directly in the pixel space. In contrast, methods like Stable Diffusion \cite{rombach2022high} and DALL-E \cite{ramesh2022hierarchical} apply the diffusion process in a low-dimensional embedding space.  Notably, Stable Diffusion has demonstrated superior image quality and efficiency \cite{zhang2023text}. Several extensions and improvements to the Stable Diffusion framework have been proposed \citep[e.g.,][]{esser2024scaling, peebles2023scalable, podell2023sdxl}. A key challenge with diffusion models, however, is their misalignment with human preferences. Recent work addresses this by controlling pre-trained models towards preferred properties, either during training or via training-free methods \cite{liu2024alignment}.

\paragraph{Training-based alignment \cite{liu2024alignment}}uses supervised fine-tuning (SFT) of the diffusion model combined with reinforcement learning from human feedback (RLHF) to align the model with human preferences, approximated via a reward model. Models in this category, such as ReFL \cite{xu2024imagereward}, DDPO \cite{black2023training}, AlignProp \cite{prabhudesai2023aligning}, DRaFT \cite{clark2023directly}, DPOK \cite{fan2024reinforcement}, and DPO-Diffusion \cite{wang2024discrete}, rely on gradient-based fine-tuning of the diffusion model. Alternatively, models can be directly optimized on preference data using methods like Diffusion-DPO \cite{wallace2024diffusion}, D3PO \cite{yang2024using}, and SPO \cite{liang2024step}. Nonetheless, those training-based alignment methods often require considerable computational resources.

\paragraph{Training-free alignment \cite{liu2024alignment}} aligns diffusion models with human preferences \textit{without the need for fine-tuning the diffusion model}. The first stream of research uses both manual and systematic approaches to prompt optimization\cite{oppenlaender2023taxonomy,wang2023reprompt}. Automatic prompt optimization methods, such as Promptist \cite{hao2024optimizing} and OPT2I \cite{manas2024improving}), leverage Large Language Models (LLMs) to refine prompts. The second stream focuses on modifying negative prompts using LLMs (e.g., DPO-Diffusion \cite{wang2024discrete}) or directly learning negative embeddings (e.g., ReNeg \cite{li2024reneg}). The third stream involves editing the initial latent state, as seen in ReNO  \cite{eyring2024reno} for one-step diffusion models. The fourth stream uses text embeddings, to which our IPGO(+) belongs, as detailed below.

\paragraph{Alignment through prompt embedding optimization} includes PEZ \cite{wen2024hard}, which aligns an image with text embeddings and prompts that reflect both the image content and style. Textual Inversion \cite{gal2022image} aligns new word tokens with novel objects or styles. TextCraftor \cite{li2024textcraftor} and TexForce \cite{chen2024enhancing} align generated images with rewards by fine-tuning the CLIP text encoder within the diffusion pipeline. Our method, IPGO, is also based on optimizing text embeddings. \textit{However}, unlike methods such as PEZ and Textual Inversion, IPGO operates without access to ground-truth images, instead leveraging abstract reward models to guide manipulation within the existing text embedding space. \textit{Moreover}, in contrast to TextCraftor and TexForce, which change the text embedding space by fine-tuning the entire text encoder, IPGO explores within the original embedding space without altering its structure.

\section{Preliminaries}
\label{sec:prelim}
\paragraph{Diffusion Models} Diffusion models are probabilistic frameworks that generate an image conditioned on a text prompt by sequentially denoising an image from pure Gaussian noise using an error model $\epsilon_\phi$ \cite{rombach2022high}, parameterized by $\phi$. The model $\epsilon_\phi$ predicts the noise in the image $x_t$, which is obtained by progressively adding Gaussian noise $\epsilon$ to the original image $x_0$ at each step of the sequence $t=0,..,T$. The model is trained by maximizing a variational lower bound, as outlined in \cite{ho2020denoising}. Text-to-image diffusion models are trained using a guidance function \cite{dhariwal2021diffusion}. For classifier-free guidance (CFG) the model is trained to predict both conditional and unconditional noise scores, which are combined during inference with a guidance scale to allow for fine-tuning of the trade-off between prompt alignment and image quality 
\cite{ho2022classifier}.

%\[
%\mathcal{L}(\phi) = \mathbb{E}_{t \sim U(0, T), x_0 \sim q_\text{data}, \epsilon \sim \mathcal{N}(0, I)} \left[ \| \epsilon - \epsilon_\phi (x_t, t) \|^2 \right].
%\]

% Let $p_0(x_0)$ be the conditional distribution of image data $x_0\in\mathbb{R}^n$. The distribution $p_0$ results from a Markovian \textit{reverse} process with Gaussian transitions between the latents $x_1,...,x_T$, according to the dynamics (see \cite{ho2020denoising}):
% \begin{equation}
% \label{eqn:diffusion_reverse}
%     p_{\theta}(x_{t-1}|x_{t})=\mathcal{N}(\mu_{\theta}(x_t,t),\Sigma_t), ~~ p(x_T)=\mathcal{N}(0,I),
% \end{equation}
% which leads to the joint distribution,
% \begin{equation}
% \label{eqn:joint_distribution}
% p_{\theta}(x_{0:T})=p(x_T)\prod_{t=1}^Tp_{\theta}(x_{t-1}|x_t).
% \end{equation}
% To apply variational inference, the posterior $q(x_{1:T}|x_0)$ is approximated by a \textit{forward} or \textit{diffusion} process, which is a Markov process that iteratively adds Gaussian noise to the given data with a variance schedule $\beta_1,\dots,\beta_T$:
% \begin{align}
%     \label{eqn:posterior_approx}
% q_{\theta}(x_{1:T}|x_0)&=\prod_{t=1}^Tq(x_t|x_{t-1}),~q(x_t|x_{t-1}) \notag \\
% &=\mathcal{N}\left(\sqrt{1-\beta_t}x_{t-1},\beta_t I\right).
% \end{align}

% The diffusion model is trained by minimizing the KL distance between the forward and reverse Gaussian Markov probabilities:
% \begin{equation}
% \label{eqn:KL_divergence}
% \mathbb{E}_q\left(\sum_{t=1}^TKL(q(x_{t-1}|x_t,x_0)||p_{\theta}(x_{t-1}|x_t))\right).
% \end{equation}

\paragraph{Reward Models} Typically, a generated image is evaluated using a pre-trained reward model, denoted as $\mathcal{S}$, which assesses how well an image produced by a diffusion model aligns with human evaluations. For each image $x$ generated by the diffusion model in response to a prompt $p$, the reward model assigns a reward $\mathcal{S}(x,p)$, serving as a proxy for human evaluation of the prompt-image pair. This reward $\mathcal{S}(x,p)$ is then used to guide the diffusion model towards generating more preferred images through optimization. To ensure the flexibility and broad applicability of our method, we consider publicly available reward models $\mathcal{S}$. Among the widely used models are the LAION aesthetic predictor V2 \citep{LaionAes}, the CLIP loss derived from the multimodal CLIP model\citep{radford2021learning}, and the human preference score HPSv2 \citep{wu2023human}. These models have played a critical role in aligning the outputs of diffusion models with human preferences in research and practice.

\section{Methods}
\label{sec:methods}

Suppose we have a prompt $p$ sampled from the prompt distribution $q_{\text{prompt}}(p)$; a well-trained reward model $S(x,p)$ on image $x$ and the prompt $p$; a text encoder $\mathcal{T}(\cdot)$ which converts $p$ to its text embeddings $\mathcal{T}(p)\in\mathbb{R}^{d\times K}$, where $d$ is the embedding dimension and $K$ is the length of the tokenized prompt; and finally a diffusion model characterized by $q_{\text{image}}(x_0|\mathcal{E},z_T)$, which represents the probability distribution of the image $x_0$ given by a set of text embeddings $\mathcal{E}$ and a fixed initial latent state $z_T$ at timestep $T$. In the following sections, we first present the base IPGO, and then introduce its extension, IPGO+, which we apply to prompt batch training.

\subsection{IPGO}
\label{subsec:IPGO}
% Optimizing prompts $p$ directly is challenging due to the discrete nature of tokens, therefore, we optimize the prompt \textit{indirectly} by optimizing its prefix and suffix in the text embedding space of the text encoder. Specifically, in diffusion models, the text prompts $C$ are first transformed into text embeddings using a text encoder $f$ (typically a transformer). These embeddings, denoted as $f(C)\in\mathbb{R}^{d\times S}$, where $d$ is the embedding dimension and $S$ is the length of the token sequence, are then fed into the diffusion model to generate the image. Importantly, embeddings $E$ are not inherently discrete, allowing us to optimize the  inputs to the diffusion model.

% \vspace{-4mm}

IPGO(+) adds to the original embeddings of a text prompt $p$ a prefix $V_{\text{pre}}$ and a suffix $V_{\text{suff}}$, consisting of $N_{\text{pre}}$ and $N_{\text{suff}}$ trainable embeddings, each of dimension $d$, and parameterized by  $\Omega_{\text{IPGO}}$ (see the following paragraphs). IPGO(+) inserts the prefix to the beginning and the suffix to the end of $\mathcal{T}(p)$, the embeddings of the prompt $p$, thereby producing a new set of text embeddings:
\begin{equation}
\label{eqn:new_embedding}
%oplus
\mathcal{E}(V_{\text{pre}},p,V_{\text{suff}};~\Omega_{\text{IPGO}}) = V_{\text{pre}}\oplus\mathcal{T}(p)\oplus V_{\text{suff}},
\end{equation}
where $\mathcal{E}(V_{\text{pre}},p,V_{\text{suff}};~\Omega_{\text{IPGO}})\in\mathbb{R}^{d\times (N_{\text{pre}}+K+N_{\text{suff}})}$ and $\oplus$ stands for the concatenation in the second dimension. 

IPGO(+) aims to optimize $\Omega_{\text{IPGO}}$ such that the expected rewards of the corresponding images sampled from $q_{\text{image}}$ conditioned on a fixed $z_T$ and $\mathcal{E}(V_{\text{pre}},p,V_{\text{suff}};~\Omega_{\text{IPGO}})$ are maximized, using the following objective function:
\begin{equation}
\label{eqn:objective_L}
   \mathcal{L}(\Omega_{\text{IPGO}}) = \mathbb{E}_{p\sim q_{\text{prompt}}(p),x_0\sim q_{\text{image}}(x_0|\mathcal{E}(V_{\text{pre}},p,V_{\text{suff}};~\Omega_{\text{IPGO}}),z_T)} \mathcal{S}(x_0,p),
\end{equation}

\begin{figure*}[!t]
    \centering
    \includegraphics[width=0.9\linewidth]{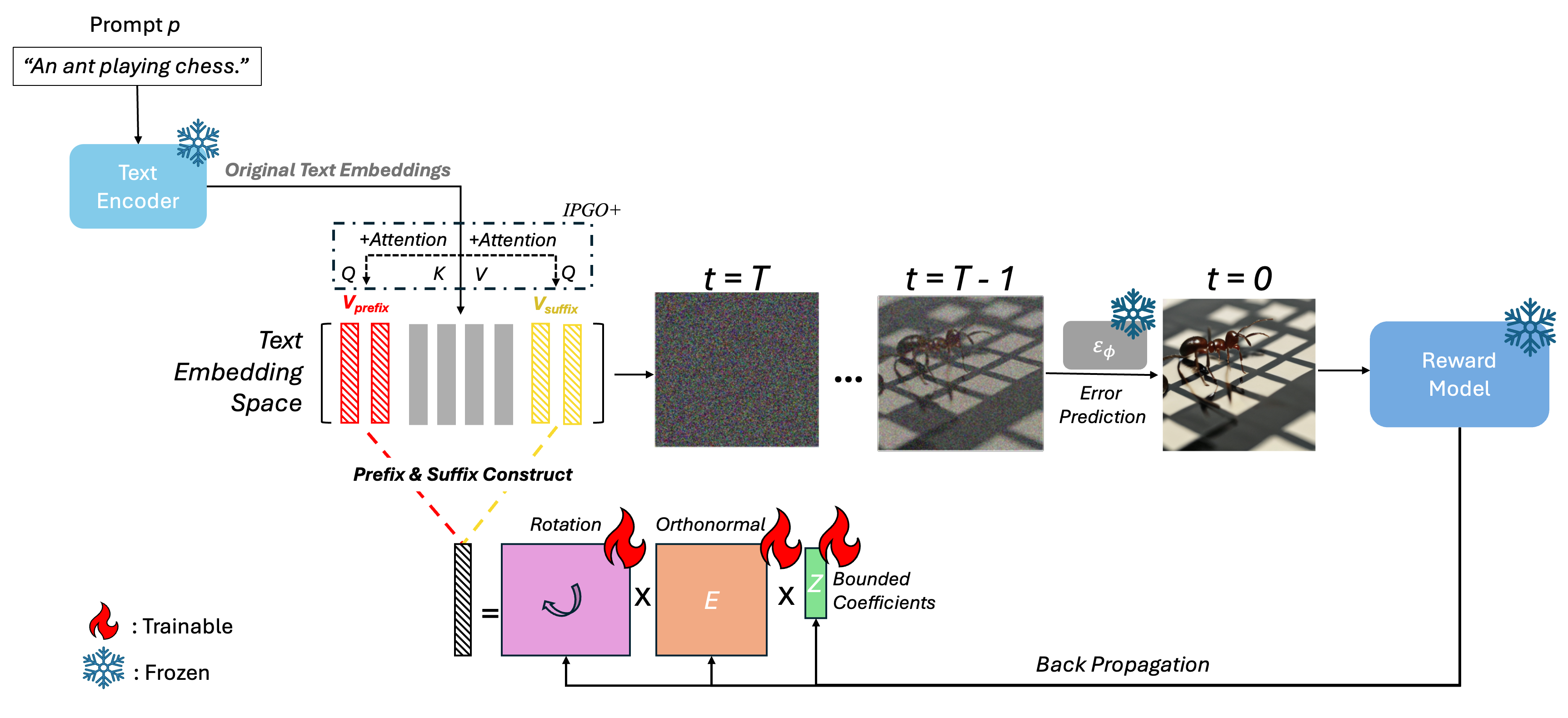}
    \vspace{-4mm}
    \caption{IPGO(+) inserts trainable prefix and suffix embeddings to the text embeddings of the prompt in the CLIP text encoder space/text embedding space, and then sends back reward signals through backpropagation under three constraints: Orthonormality, Range and Conformity. IPGO+ further  adds an additional attention layer (indicated by the dashed black box) to the text embeddings prior to image sampling.}
    \label{fig:IPGO_diagram}
\end{figure*}

In the following sections, we present the motivation behind our approach and outline the overall framework. Figure \ref{fig:IPGO_diagram} illustrates the overview of the IPGO(+) methodology. %, then extend it to a Bayesian Optimization scheme.

\paragraph{Constrained Prefix-Suffix Tuning} Inspired by Prefix-Tuning \citep{li2021prefix}, we propose IPGO(+) for aligning diffusion models, which involves adding extra continuously differentiable embeddings before and after the original text embeddings, as described by equation \ref{eqn:new_embedding}. Li and Liang \citep{li2021prefix} show that directly updating prefix embeddings may lead to unstable optimization. Thus, rather than directly optimizing $V_{\text{pre}}$ and $V_{\text{suff}}$, we reparameterize them as rotated linear combinations of a set of low-dimensional learnable embeddings, subject to three sets of constraints. We parameterize $V_{*}$ ($*$ stands for "pre" or "suff" from here on) by the following:
\begin{align}
\label{eqn:v_pre and v_suf}
    V_{*} = \tilde{R}_{2,\theta_2^{*}}\tilde{R}_{1,\theta_1^{*}}E_{*}Z_{*},
\end{align}
where $E_{*}\in\mathbb{R}^{d\times m_{*}}$ is a trainable set of \textit{base text embeddings} and $m_{*}$ is the length of the basis. $Z_{*}\in\mathbb{R}^{N_{*}\times m_{*}}$ are linear coefficients for the basis. In words, the basis helps explore a subspace of the text embedding space that will be optimized to fit with the reward guidance. Exploring the subspace is more efficient than exploring the original embeddings. In addition to the linear parameterization, we apply two rotation matrices $\tilde{R}_{1,\theta_1^{*}}$ and $\tilde{R}_{2,\theta_2^{*}}$. The rotation matrices are composed of the $2\times2$ elementary matrix controlled by a given angle $\theta\in(-\frac{\pi}{2},\frac{\pi}{2}]$:
\begin{equation}
\label{eqn:rotation_element}
R_{e,\theta} = \begin{bmatrix}
\cos{\theta} & -\sin{\theta} \\
\sin{\theta} & \cos{\theta}
\end{bmatrix}.
\end{equation}
We constrain $\theta\in(-\frac{\pi}{2},\frac{\pi}{2}]$ because we are rotating the 1-dimensional line of the text embedding. Now we define the rotation matrices in Equation \ref{eqn:v_pre and v_suf}. Given two angles $\theta_1^*$ and $\theta_2^*$, we define $\tilde{R}_{1,\theta_1^*}\in\mathbb{R}^{d\times d}$ and $\tilde{R}_{2,\theta_2^*}\in\mathbb{R}^{d\times d}$ by:
\begin{align}
\label{eqn:rotation_matrices_defn}
    \tilde{R}_{1,\theta_1^*}&=I_{d/2}\otimes R_{e,\theta_1^*} \\
    \tilde{R}_{2,\theta_2^*}&=\begin{bmatrix}
        R_{e,\theta_2^*,(2)} & & \\
        & I_{d/2-1}\otimes R_{e,\theta_2^*} & \\
        & &R_{e,\theta_2^*,(1)}
    \end{bmatrix},
\end{align}

where $\otimes$ is the tensor product, $I_a$ is the identity matrix of size $a$, $R_{e,\theta_2^*,(i)}$ is the $i^{th}$ row of the elementary rotation matrix $R_{e,\theta_2^*}$. To interpret, $\tilde{R}_{1,\theta_1^*}$ rotates $(2j-1,2j)$ pairs, while $\tilde{R}_{2,\theta_2^*}$ rotates $(2j,2j+1)$ pairs of coordinates of a given right-multiplied embedding, where $j=1,\dots,d/2$ and the $(d+1)^{th}$ coordinate is the $1^{st}$ coordinate.
Rotations are advantageous in two ways. First, they introduce a certain amount of non-linearity via the additional angle parameters, which increases the search space. Second, they accelerate the search process by modifying the gradient directions (see the details in Appendix \ref{appx_sec: rotation}). 

To maximize the subspace and preserve the structural integrity of the text embeddings generated by the text encoder $\mathcal{T}$, three additional constraints are imposed. First, we impose an \textit{\textbf{Orthonormality Constraint}} on the text embedding basis $E_*$, i.e.,
\begin{equation}
\label{eqn:orthonormal_constraint}
E_{*}E_{*}^T=I_{m_*}.
\end{equation}
Orthonormality standardizes the subspace basis and ensures maximal subspace size.

Second, to further reduce the likelihood of exploring extreme text embeddings that harm the original prompt, we introduce a \textit{\textbf{Range Constraint}}, which restricts the values of the affine transformation coefficients $Z_*$ to the range $[-1,1]$. Concretely, the lengths of the prefix and suffix embeddings are controlled so that they will not perturb the original semantics. 

Third, we add a \textit{\textbf{Conformity Constraint}} to ensure that the embeddings of the inserted embeddings align closely with those of the original prompt statistically, promoting consistency and coherence of the generated prefix and suffix embeddings with the original prompt:
\begin{equation}
\label{eqn:sum constraint}
% \frac{1}{N_{\text{pre}}+K+N_{\text{suff}}}\left(\sum_{i=1}^{N_{\text{pre}}}V_{\text{pre},i}+\sum_{i=1}^{K}\mathcal{T}(p)_{i}+\sum_{i=1}^{N_{\text{suff}}}V_{\text{suff},i}\right) = \frac{1}{K}\sum_{j=1}^{K}\mathcal{T}(p)_{j},
\text{Mean}(\mathcal{E}(V_{\text{pre}},p,V_{\text{suff}};~\Omega_{\text{IPGO}}))=\text{Mean}(\mathcal{T}(p)),
\end{equation}

where Mean($\cdot$) calculates the average of the input embeddings. In words, the semantics of the new text embeddings should remain roughly the same as the original one.

In summary, IPGO has parameters $\Omega_{\text{IPGO}}=\{E_{\text{pre}}, E_{\text{suff}},\theta_1^{\text{pre}}, \theta_2^{\text{pre}}, \theta_1^{\text{suff}},\theta_2^{\text{suff}},Z_{\text{pre}}, Z_{\text{suff}}\}$, optimized by the following objective:
\begin{equation}
\label{eqn:optimization}
\begin{aligned}
\max_{\Omega_{\text{IPGO}}} \quad & \mathbb{E}_{p\sim q_{\text{prompt}}(p),x_0\sim q_{\text{image}}(x_0|\mathcal{E}(V_{\text{pre}},p,V_{\text{suff}};\Omega_{\text{IPGO}}),z_T)} \mathcal{S}(x_0,p)\\
\textrm{s.t.} \quad & (\ref{eqn:orthonormal_constraint}),(\ref{eqn:sum constraint}), Z_{\text{pre},ij}\in[-1,1], Z_{\text{suff},ij}\in[-1,1]\\
&(\theta_1^{\text{pre}},\theta_2^{\text{pre}})\in \left(-\frac{\pi}{2},\frac{\pi}{2}\right]^2, (\theta_1^{\text{suff}},\theta_2^{\text{suff}})\in \left(-\frac{\pi}{2},\frac{\pi}{2}\right]^2,
\end{aligned}
\end{equation}
where $Z_{*.ij}$ is the $(i,j)$ entry of the matrix $Z_*$.

%\paragraph{Remarks:} Notice that our formulation for $V_*$'s is related to LoRA \citep{hu2021lora} and RoPE \citep{su2024roformer}, as both of these parameterizations exploit a low-rank structure and rotations. But our formulation for the prompt optimization context has stronger constraints and operates entirely at the prompt-level instead of during model fine-tuning. Our rotation operations, unlike RoPE, are not designed to encode positions, but to add non-linearity and enlarge the search space by including trainable angle parameters.

\subsection{IPGO+}
\label{subsec:IPGO+}
\paragraph{Attention-aware Prefix and Suffix for Prompt-Batch Training (IPGO+)} To facilitate prompt-batch training, we introduce a layer with a cross-attention mechanism on the prefix and suffix embeddings (Queries: $Q$) and the original prompt embeddings (Key: $K$ and Value: $W$) to reinforce the interactions between the original prompt and the inserted embeddings (see the dashed black box in Figure \ref{fig:IPGO_diagram}). This layer is purely computational and does \textbf{NOT} involve any additional parameters. We use the standard scaled dot attention, as implemented in PyTorch\footnote{\url{https://pytorch.org/docs/stable/generated/torch.nn.functional.scaled_dot_product_attention.html}}, which is represented by \cite{vaswani2017attention}: 
%\vspace{-2mm}
%\begin{equation}
%    \label{eqn:scaled_attention}
$ \text{Attention}(Q,K,W)=\text{softmax}\left(\frac{QK'}{\sqrt{d}}\right)W.$
%\end{equation}
With this, the attention-aware prefix and suffix $V_{*}^{\text{att}}$ are of the form:
\begin{equation}
    \label{eqn:attention_prefix_suffix}
    V_{*}^{\text{att}} = V_{*} + \text{Attention}(V_{*},\mathcal{T}(p),\mathcal{T}(p)).
\end{equation}

Therefore, IPGO+ has the following objective function with the same parameter space $\Omega_{\text{IPGO}}$:
\begin{equation}
\label{eqn:objective_L_IPGO+}
   \mathcal{L}^{(+)}(\Omega_{\text{IPGO}}) = \mathbb{E}_{p\sim q_{\text{prompt}}(p),x_0\sim q_{\text{image}}(x_0|\mathcal{E}(V_{\text{pre}}^{\text{att}},p,V_{\text{suff}}^{\text{att}};~\Omega_{\text{IPGO}}),z_T)} \mathcal{S}(x_0,p).
\end{equation}

\section{Experiments and Results}
\label{sec:experiments}
We conduct a comprehensive set of experiments to evaluate the performance of IPGO(+) across seven benchmark models on three datasets. This section is structured as follows: In Section \ref{exp-setting}, we describe the experiment settings, while Section \ref{subsec:benchmarks} introduces the benchmarks. Section \ref{subsec:promptwise-training} presents evaluations and comparisons with baselines for prompt-wise optimization using the base IPGO. Results of prompt-batch training with IPGO+ are detailed in Section \ref{subsec:prompt-batch}. Finally, Appendix \ref{appx:abalation} provides detailed analyses and ablation studies.

\subsection{Experiment Settings}
\label{exp-setting}
\paragraph{Datasets.} Three datasets are considered: the COCO image captions \cite{lin2014microsoft}, DiffusionDB \cite{wang2022diffusiondb}, and Pick-a-Pic \cite{kirstain2023pick}. These datasets represent a wide range of prompts and images with varying levels of complexity. To assess the performance of IPGO(+) across different categories of image captions, we conduct separate evaluations for COCO images in the following categories: Persons, Rooms, Vehicles, Natural Scenes, and Buildings. For each category, we randomly select 60 captions from the COCO dataset. Additionally, we randomly select 300 prompts from both the DiffusionDB and Pick-a-Pic datasets, resulting in a total of 900 prompts for evaluation. This selection size exceeds those used in recent experiments, such as those by \cite{black2023training} and \cite{wang2024discrete}, which both used about 600 prompts.

\paragraph{Training.} All experiments with IPGO(+), which has a total of 0.47M parameters, are conducted on a single NVIDIA L4 GPU with 22.5GB of memory. IPGO(+) takes at most 12GB of memory for all tasks. The backbone diffusion model used is Stable-Diffusion v-1.5, chosen for its better balance between  generation quality and computational efficiency \citep{li2024textcraftor, podell2023sdxl}. Note that IPGO(+) is directly applicable  to other diffusion models (Appendix \ref{appx:adaptability_IPGO}). For consistency, all models and experiments are run in identical computational environments. In contrast, the benchmark TextCraftor (introduced below) requires a single A100 GPU.

%because it has been shown to perform well in real-world human evaluations with attractive model size and computation as compared to other diffusion models. A DDIM Scheduler is employed and the generated images have a resolution of $512\times512$ pixels. During optimization, we truncate the backpropagation at the second last sampling step. We also apply gradient clipping across all our experiments, selecting a gradient clipping norm of $c=1.0$.

% To ensure consistency, we establish the identical computational environments across all models and experiments, which allows each full training run on a dataset to complete within one day on a single GPU. For the benchmark TextCraftor (introduced below), however, a single A100 GPU is used. 

\paragraph{Reward Models.} We evaluate performance using three publicly available reward models that focus on image aesthetics, image-text alignment, and human preference. Specifically, we use the LAION aesthetic predictor V2 \citep{LaionAes}, the CLIP loss from the multimodal CLIP model \citep{radford2021learning}, and the human preference score v2 \citep{wu2023human}. These widely used reward models capture a broad spectrum of criteria, effectively representing the diverse rewards relevant to text-image alignment tasks. 

\vspace{-2mm}
\subsection{Benchmarks}
\label{subsec:benchmarks}
We evaluate IPGO using the following benchmarks, which represent the current state of the art (SOTA) in the categories discussed in Section~\ref{sec:related_work}. The first baseline is \textbf{Stable diffusion with a raw prompt} \cite{rombach2022high}, against which we expect IPGO(+) to enhance performance across all datasets and reward models. The second baseline is \textbf{TextCraftor}, using a fine-tunable text encoder with 123M parameters, representing the current SOTA among text-embedding-based methods. We also use two training-based methods: \textbf{DRaFT} \cite{clark2023directly} and \textbf{DDPO} \cite{black2023training}. For DRaFT, we select the DRaFT-1 variant with LoRA of rank 3 as the baseline (\#parameters: 0.60M), due to its low computational cost and competitive performance \cite{clark2023directly}. For DDPO (\#parameters: 0.79M), we apply the default LoRA configuration. Furthermore, we include three training-free methods: \textbf{DPO-Diff} \cite{wang2024discrete}, \textbf{Promptist} \cite{hao2024optimizing}, and \citep[following the idea of][]{manas2024improving} \textbf{ChatGPT-4o}. For ChatGPT-4o, we prompt with the following instruction: \textit{Improve this sentence to ensure that its stable-diffusion-generated image is \{reward-based property\}, while remaining concise.} Here, the \{reward-based property\} is specified as ``more aesthetically pleasing," ``more aligned with human preference," or ``more aligned with the original sentence," depending on the target reward. Detailed qualitative comparisons between several benchmarks and IPGO can be found in Appendix \ref{appx:qual_compare}, while the hyperparameter settings are provided in Table \ref{tab:hyperparameters} in Appendix \ref{appx:hyperparameters}.

\subsection{Prompt-Level Image Optimization at Inference}
\label{subsec:promptwise-training}
We first perform prompt-wise training for all methods listed in Table \ref{tab:qualitative_compare}, where each optimization uses a single prompt. Single image optimization during inference is more flexible and addresses concerns regarding generalization to unseen prompts \cite{eyring2024reno}. Here we use the base IPGO without the cross-attention mechanism to establish a performance baseline. All seven benchmarks are evaluated using default configurations for learning and sampling. The best loss value achieved during training is used to represent the final performance of each method.

Table \ref{tab:individual_training_clip_alignment} presents the results of IPGO and benchmark methods for semantic alignment across three datasets. The highest scores are highlighted in bold. With CLIP alignment loss, IPGO outperforms all seven benchmarks in all scenarios except for COCO-Building. Nevertheless, IPGO achieves the highest average alignment scores across all prompts, surpassing the top benchmarks, TextCraftor and DRaFT-1, by 2.4\% and 2.7\%, respectively. %Promptist and ChatGPT-4o offer minimal gains, likely due to limited generalization to unseen prompts. %only TextCraftor and DRaFT-1 are competitive with IPGO.  They outperform IPGO on the COCO dataset containing images with buildings; in all other scenarios, IPGO surpasses these two best-performing benchmark models; clearly its average reward score across all datasets is higher than that of TextCraftor and DRaFT-1, which ar on par. IPGO improves over TextCraftor by an average of 2.4\%, and over DRaFT-1 by an average of 2.7\%. While DPO-Diff also achieves strong results, the discrete search space limits its performance relative to IPGO. Promptist and ChatGPT-4o show minimal improvements, likely due to their limited generalizability to new prompts.
\begin{table}[!htb]
\resizebox{\textwidth}{!}{\begin{tabular}{lcccccccc}
\hline
 \textit{Dataset} & \textit{IPGO} & \textit{SD v1.5} & \textit{TextCraftor} & \textit{DRaFT-1} & \textit{DDPO} & \textit{DPO-Diff} & \textit{Promptist} & \textit{ChatGPT 4o} \\
\hline
\textit{COCO} \\
\quad \textit{Person}         & \textbf{0.3160} & 0.2637 & \underline{0.3085} & 0.3067 & 0.2865 & 0.2911   & 0.2598    & 0.2581     \\
\quad \textit{Room}           & \textbf{0.2883} & 0.2482 & \underline{0.2786} & 0.2782 & 0.2648 & 0.2755   & 0.2398    & 0.2419     \\
\quad \textit{Vehicle}        & \textbf{0.2986} & 0.2514 & 0.2928 & \underline{0.2934} & 0.2755 & 0.2881   & 0.2474    & 0.2467     \\
\quad \textit{Natural Scenes} & \textbf{0.2922} & 0.2539 & \underline{0.2876} & 0.2802 & 0.2614 & 0.2815   & 0.2307    & 0.2357     \\
\quad \textit{Buildings}      & 0.2846 & 0.2439 & \underline{0.2859} & \textbf{0.2898} & 0.2718 & 0.2794   & 0.2377    & 0.2401     \\
\textit{DiffusionDB}   & \textbf{0.3247} & 0.2759 & 0.3146& \underline{0.3167} & 0.3024 & 0.2929   & 0.2753    & 0.2702     \\
\textit{Pick-a-Pic}     & \textbf{0.3125} & 0.2681 & 0.3077 & \underline{0.3103} & 0.2946 & 0.2980   & 0.2612    & 0.2595  \\  
\midrule
 \textit{Avg. Reward} & \textbf{0.3024}	& 0.2579 & \underline{0.2965} & \underline{0.2965}	& 0.2796 & 0.2866 & 0.2503& 0.2502 \\
\hline
\end{tabular}}
\vspace{-1mm}
\caption{Comparison of IPGO’s \textbf{alignment scores} with seven benchmarks across 900 prompts from three datasets. Bold/underline denote highest/second-highest scores.}
\vspace{-6mm}
\label{tab:individual_training_clip_alignment}
\end{table}

Table \ref{tab:individual_training_clip_aesthetics} presents the results for LAION aesthetics scores. IPGO outperforms all benchmarks across every dataset. Its average reward score across is the highest across all datasets, with a notable improvement of 3.2\% over the best-performing benchmark, TextCraftor. %In contrast, DPO-Diff degrades the aesthetics scores.  
\begin{table}[!htb]
\resizebox{\textwidth}{!}{\begin{tabular}{lcccccccc}
\hline
 \textit{Dataset} & \textit{IPGO} &  \textit{SD v1.5} & \textit{TextCraftor} & \textit{DRaFT} & \textit{DDPO} & \textit{DPO-Diff} & \textit{Promptist} & \textit{ChatGPT 4o} \\
\hline
\textit{COCO}\\
\quad \textit{Person}         & \textbf{6.2174} & 5.2447 & 5.8365 & 5.7761 & 5.5777 & 4.2865   & \underline{5.9401}    & 5.2297     \\
\quad \textit{Room}           & \textbf{5.7549} & 5.0931 & \underline{5.5994} & 5.4426 & 5.3700 & 4.1589   & 5.5993    & 5.2533     \\
\quad \textit{Vehicle}        & \textbf{5.8567} & 4.9608 & \underline{5.6699} & 5.5063 & 5.4219 & 4.0197   & 5.5643    & 5.0654     \\
\quad \textit{Natural Scenes} & \textbf{5.9301} & 5.0558 & \underline{5.7436} & 5.6156 & 5.5099 & 4.2952   & 5.6483    & 5.1536     \\
\quad \textit{Buildings}      & \textbf{5.7987} & 5.0326 & \underline{5.6909} & 5.4294 & 5.3484 & 4.2777   & 5.6431    & 5.2059     \\
\textit{DiffusionDB}   & \textbf{6.3469} & 5.5012 & \underline{6.3297} & 6.1100 & 5.9644 & 4.4350   & 5.6291    & 5.5878     \\
\textit{Pick-a-Pic}     & \textbf{6.2684} & 5.3289 & \underline{6.0120} & 5.9048 & 5.7547 & 4.3565   & 5.6954    & 5.4103 \\ 
\midrule
 \textit{Avg. Reward} & \textbf{6.0247}	& 5.1739 & \underline{5.8403} & 5.6835 & 5.5639 & 4.2614 & 5.6742 &	5.2708 \\
\hline
\end{tabular}}
\vspace{-1mm}
\caption{Comparison of IPGO’s \textbf{aesthetics scores} with seven benchmarks across 900 prompts from three datasets. Bold/underline denote highest/second-highest scores.}
\vspace{-6mm}
\label{tab:individual_training_clip_aesthetics}
\end{table}

Table \ref{tab:individual_training_clip_preferences} presents the results of IPGO and benchmark methods for human preference scores. Again, IPGO outperforms all benchmarks in all datasets. Its average reward score across all datasets is the highest, achieving an average improvement of 1.0\% over the strongest benchmark TextCraftor. While most other benchmark models struggle to improve the raw prompt to achieve better preference scores, IPGO consistently improves performance across all COCO categories and datasets. 

\begin{table}[!htb]
\resizebox{\textwidth}{!}{\begin{tabular}{lcccccccc}
\hline
\textit{Dataset} & \textit{IPGO} & \textit{SD v1.5} & \textit{TextCraftor} & \textit{DRaFT-1} & \textit{DDPO} & \textit{DPO-Diff} & \textit{Promptist} & \textit{ChatGPT 4o} \\
\hline
\textit{COCO}\\ 
\quad \textit{Person}           & \textbf{0.2950} & 0.2796 & \underline{0.2905} & 0.2786 & 0.2819 & 0.2481   & 0.2680    & 0.2739     \\
\quad \textit{Room}             & \textbf{0.2847} & 0.2673 & \underline{0.2806} & 0.2646 & 0.2711 & 0.2430   & 0.2596    & 0.2671     \\
\quad \textit{Vehicle}          & \textbf{0.2930} & 0.2761 & \underline{0.2905} & 0.2755 & 0.2814 & 0.2491   & 0.2679    & 0.2728     \\
\quad \textit{Natural Scenes}   & \textbf{0.2890} & 0.2721 & \underline{0.2848} & 0.2667 & 0.2741 & 0.2487   & 0.2600    & 0.2687     \\
\quad \textit{Buildings}        & \textbf{0.2916} & 0.2719 & \underline{0.2882} & 0.2723 & 0.2782 & 0.2580   & 0.2634    & 0.2697     \\
\textit{DiffusionDB}     & \textbf{0.2729} & 0.2594 & \underline{0.2719} & 0.2602 & 0.2634 & 0.2381   & 0.2585    & 0.2573     \\
\textit{Pick-a-Pic}       &\textbf{0.2753} & 0.2621 & \underline{0.2741} & 0.2647 & 0.2672 & 0.2509   & 0.2591    & 0.2619    \\
\midrule
 \textit{Avg. Reward} & \textbf{0.2859}	& 0.2698& \underline{0.2829} & 0.2689	& 0.2739 & 0.2480 & 0.2624 &	0.2673\\
\hline
\end{tabular}}
\vspace{-1mm}
\caption{Comparison of IPGO’s \textbf{human preference scores} with seven benchmarks across 900 prompts from three datasets. Bold/underline denote highest/second-highest scores.}
\vspace{-5mm}
\label{tab:individual_training_clip_preferences}
\end{table}

\begin{figure*}[!h]
\vspace{-3mm}
    \centering
    \includegraphics[width=0.85\linewidth, height=0.65\linewidth]{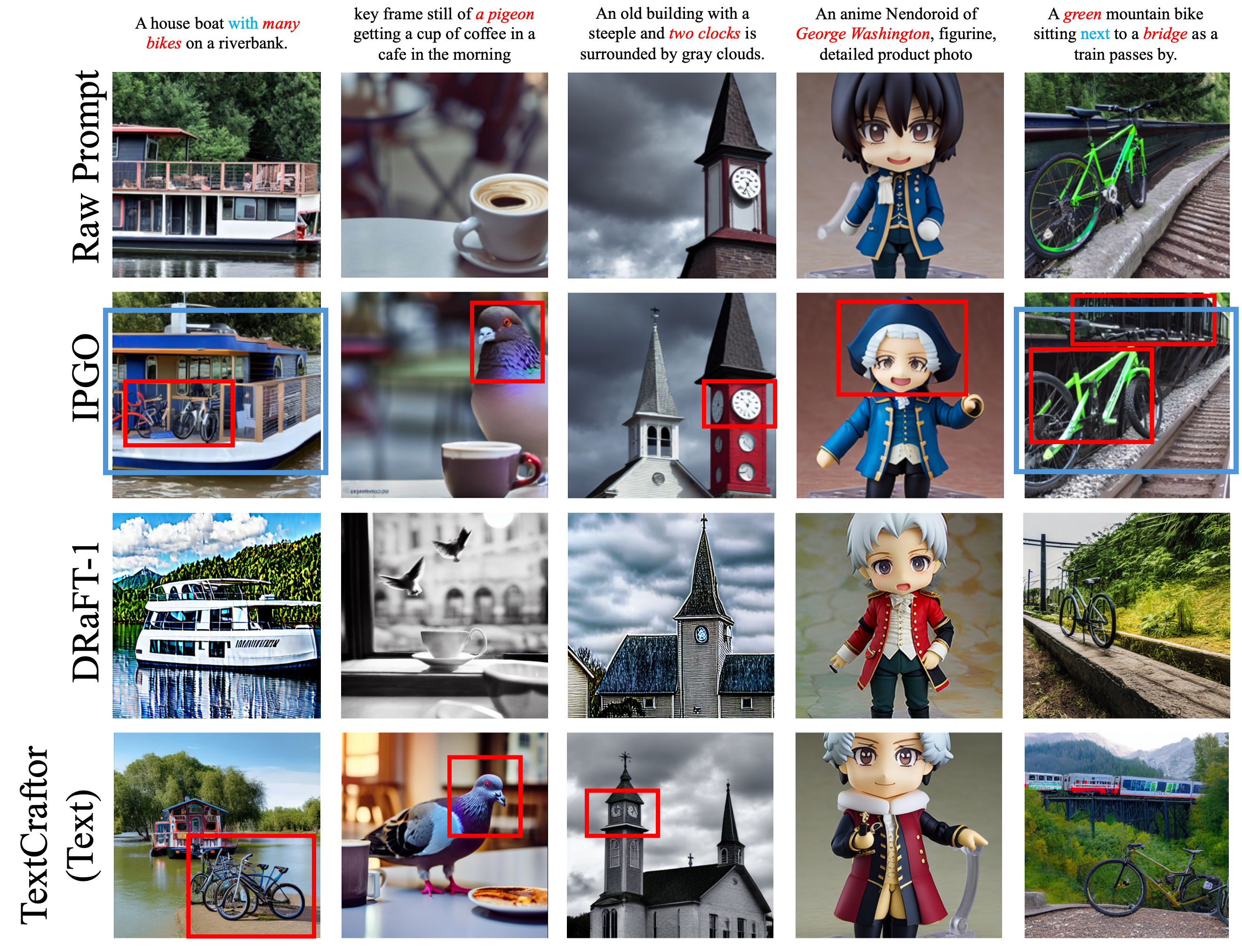}
    \vspace{-4mm}
    \caption{Example images generated with Stable Diffusion v1.5 using the raw prompt (row 1), IPGO (row 2), DRaFT-1 (row 3) and TextCraftor (row 4), evaluated according to the HPSv2 reward.}
    \label{fig:demo_promptwise_train}
\end{figure*}

Figure \ref{fig:demo_promptwise_train} qualitatively compares a sample of images generated with IPGO using the HPSv2 reward to those generated with the raw prompt. These are compared with images generated by TextCraftor and DRaFT-1, the best performing benchmarks. The figure shows that IPGO generates images that are at least on par with DRaFT-1 and TextCraftor, often surpassing them in terms of alignment of details. Unlike DRaFT-1 and TextCraftor, which may alter the image layout from that produced by the raw prompt, IPGO tends to modify or add details, while preserving the original layout. Additional examples can be found in Appendices \ref{appx:IPGO_extra_images} and \ref{appx:IPGO_more_images_others}. Note that the computational environment affects the quality of each image in Figure \ref{fig:demo_promptwise_train} equally.

These findings highlight IPGO's consistently strong performance in improving image quality with prompt-wise training at inference. IPGO achieves a substantial 1-3\% improvement over the best-performing benchmarks, which is similar to improvements reported for prior models \citep[e.g.,][]{black2023training,clark2023directly, hao2024optimizing, li2024textcraftor}. Across all 147 comparisons with seven benchmarks, IPGO has an \textit{impressive win rate of 99\%}.
% These findings indicate that IPGO performs very well across datasets and image categories by improving single prompts to achieve superior text-image alignment, human preference, and especially aesthetics scores. The improvements of IPGO over the best-performing benchmarks of 1-3\% on average are substantial, systematic, and comparable in magnitude to prior reports in the literature \citep[e.g.,][]{black2023training,clark2023directly, hao2024optimizing, li2024textcraftor}. Across the 147 comparisons, IPGO has an \textit{impressive win rate of 99\%} over the seven benchmarks.

\vspace{-5mm}
\subsection{Prompt-Batch Prefix-Suffix Learning with IPGO+}
\label{subsec:prompt-batch}
% We evaluate IPGO+'s performance in prompt-batch training, where its prefix and suffix tokens are shared across all prompts in a batch, rather than being customized for each individual prompt. We use IPGO+ which includes the attention mechanism described in Equation~\ref{eqn:attention_prefix_suffix}, because it allows shared tokens to act as localized adjustments to the individual prompt text embeddings. To assess the impact of prompt heterogeneity within batches, we conduct experiments using two configurations: (1) 60 prompts from the COCO-Person dataset, and (2) a mixed batch of 30 prompts from COCO-Person and 30 from COCO-Natural Scenes. In addition, we evaluate IPGO+ on the DiffusionDB and PickaPic datasets. Given TextCraftor's and DRaFT-1’s strong performances in Section~\ref{subsec:promptwise-training} and their suitability for batch optimization, we use them as the benchmarks. We analyze the first 1,200 image generations during training over 20 epochs, using batch sizes of 4 and 10 (corresponding to 15 and 6 batches per epoch, respectively) to study batch size effects. For faster inference and more efficient training with IPGO+, we set 30 sampling steps for the diffusion model and truncate the gradient at the tenth to the last sampling step with gradient checkpoints. Default configurations are used for the two benchmark models. Results are summarized in Table \ref{tab:batch_training}.
We evaluate IPGO+ in prompt-batch training, where shared prefix and suffix embeddings are applied across prompts in the batch. IPGO+ includes the attention mechanism in Equation~\ref{eqn:attention_prefix_suffix}, enabling localized adjustments to individual prompt embeddings. We assess the effect of prompt heterogeneity within batches by: (1) 60 COCO-Person prompts and (2) a mixed batch of 30 COCO-Person and 30 COCO-Natural Scenes prompts. We also evaluate on DiffusionDB and Pick-a-Pic. Given the strong performance of TextCraftor and DRaFT-1 in Section~\ref{subsec:promptwise-training} and their suitability for batch optimization, they serve as our benchmarks. We train for 20 epochs with batch sizes of 4 and 10 to study the batch-size effect. We report the final batch-wise average rewards. Training details are in Appendix \ref{appx:hyperparameters}. The results are in Table~\ref{tab:batch_training}. Example images and comparisons with TextCraftor are in Appendix \ref{appx:IPGO+_extra_images}.

First, on average, across batch sizes and datasets, IPGO+ achieves the highest CLIP alignment scores. For the COCO-Person, COCO-Mix, and DiffusionDB datasets, IPGO+ outperforms the second-best model, TextCraftor, for both batch sizes. However, IPGO+ performs slightly worse for the COCO mixed dataset compared to COCO-Person. For the Pick-a-Pic dataset, TextCraftor outperforms IPGO+ and DRaFT-1 in all cases. These findings suggest that while IPGO+ demonstrates strong performance, %it does not always outperform TextCraftor in learning tokens that ensure coherent semantic structures, even with the attention mechanism. 
finding common prefix and suffix embeddings that are semantically consistent with all prompts is challenging for IPGO, especially as the batch size increases ($B=10$). In fact, the average CLIP alignment scores for prompt-level optimization on the COCO-person, DiffusionDB and Pick-a-Pic datasets, shown in Table \ref{tab:individual_training_clip_alignment}, are higher than the scores achieved in batch optimization ($B=4$ and $B=10$) in Table \ref{tab:batch_training}.

Second, for aesthetics rewards, IPGO+ outperforms both TextCraftor and DRaFT-1 on average, yielding a 2.9\% improvement over the strongest benchmark, TextCraftor.  IPGO+ outperforms both TextCraftor and DRaFT-1 for the smaller batch size ($B=4$) across all datasets except Pick-a-Pic, and it tends to perform better with smaller batches. At the larger batch size ($B=10$), the results are mixed: IPGO+ performs best on the COCO-Mixed and Pick-a-Pic datasets, while DRaFT-1 achieves highest scores on COCO-Person and DiffusionDB. Note that the aesthetics scores achieved by IPGO+ in batch optimization (shown in Table \ref{tab:batch_training}) are higher than those for prompt-level optimization on the COCO-person, DiffusionDB and Pick-a-Pic datasets (shown in Table \ref{tab:individual_training_clip_aesthetics}).  

Third, IPGO+ achieves equal or better human preference scores than TextCraftor and DRaFT-1 in all cases except one (TextCraftor on the Pick-a-Pic with a batch size of 4). Across all datasets and batch sizes, IPGO+ shows an average improvement of 2.0\% over TextCraftor. TextCraftor outperforms DRaFT-1 in all cases. Again, IPGO+ tends to perform better with smaller batch sizes ($B=4$) for COCO-Person and COCO-Mixed prompts, as well as DiffusionDB. It also performs better with homogeneous batches of prompts (COCO Person vs. COCO Mixed). Note that the average HPSv2 scores for for prompt-level optimization on the COCO-person, DiffusionDB and Pick-a-Pic datasets, as shown in Table \ref{tab:individual_training_clip_preferences}, are somewhat higher than the scores achieved in batch optimization ($B=4$ and $B=10$) in Table \ref{tab:batch_training}.

In summary, IPGO+ performs well across all three rewards relative to TextCraftor and DRaFT-1: IPGO+ achieves the highest scores in 18 out of 24 scenarios (75.0\%), and second-highest scores in 5 additional scenarios. Thus, across all datasets and batch sizes, IPGO+ consistently outperforms the benchmark models, particularly for smaller, homogeneous batches. While IPGO+ performs well in aesthetics and human preferences, for CLIP alignment (and to a lesser extent HPSv2), training IPGO+ with individual prompts may be preferred.
\begin{table}[!htb]
\vspace{-2mm}
\resizebox{\textwidth}{!}{\begin{tabular}{llccc|ccc|ccc}
\toprule
 &  & \multicolumn{3}{c}
                 {\textit{Alignment}} & \multicolumn{3}{c}      {\textit{Aesthetics}} & \multicolumn{3}{c}  {\textit{Human Preference}} \\
%\hline
\textit{Dataset}  &  \textit{Batch Size}  & \textit{IPGO+}       & \textit{TextCraftor} & \textit{DRaFT-1}  & \textit{IPGO+}      & \textit{TextCraftor} & \textit{DRaFT-1}  & \textit{IPGO+}             & \textit{TextCraftor} & \textit{DRaFT-1}  \\
\hline
COCO & & & & & & & &\\
 \quad Person       &  $B=4$  &  \textbf{0.2885}    & \underline{0.2861} & 0.1583 & \textbf{7.4127} & \underline{7.2389}   &  4.4503  & \textbf{0.2935}   & \underline{0.2794} & 0.2539 \\
\quad Person            & $B=10$ & \textbf{0.2890}    & \underline{0.2843} & 0.2742 & \underline{7.4881} & 7.1530   & \textbf{7.8473} & \textbf{0.2866}   & \underline{0.2768}  & 0.2761 \\
\quad Mixed        &  $B=4$   & \textbf{0.2817}    & \underline{0.2794} & 0.2621 & \textbf{7.8554} & \underline{7.2389}     & 5.8729   & \textbf{0.2881}  & \underline{0.2790} & 0.2708 \\
\quad Mixed  &  $B=10$  & \textbf{0.2770} & \underline{0.2766}  & 0.2695 & \textbf{7.1896} & \underline{6.7600} & 5.6144  & \textbf{0.2761} & \underline{0.2760} & 0.2654 \\
%\midrule
DiffusionDB & $B=4$ & \textbf{0.2917} & \underline{0.2906} & 0.1358 & \textbf{7.8784} & \underline{7.8715} & 7.3916  & \textbf{0.2664} & \underline{0.2572} & 0.2074\\
            & $B=10$          & \textbf{0.2846} & \underline{0.2838}    & 0.2799 & 7.6428 &  \underline{7.7298}  &  \textbf{7.9151}  & \textbf{0.2586} & \underline{0.2571}  & 0.2484\\
%\midrule
Pick-a-Pic    & $B=4$ & \underline{0.2810} & \textbf{0.2835} & 0.1344 & \underline{7.7038} & \textbf{7.8541} & 6.9854  & \underline{0.2610} & \textbf{0.2632} & 0.2239\\
            & $B=10$          & \underline{0.2780}    & \textbf{0.2803} & 0.2753 & \textbf{7.5418} & 6.4437  & \underline{7.5006}  & \textbf{0.2620} & \textbf{0.2620}   & 0.2452 \\
\midrule
 \textit{Avg. Rewards} & $B=4$ & \textbf{0.2857} & 
 \underline{0.2849} & 0.1727 & \textbf{7.7126} & \underline{7.5509} & 6.1751  & \textbf{0.2773} & \underline{0.2697} & 0.2390\\
 & $B=10$ & \textbf{0.2822} & \underline{0.2813} & 0.2747 & \textbf{7.4656} & \underline{7.0216} & 7.2194  & \textbf{0.2708} & 
 \underline{0.2680} & 0.2588\\
 & \textit{Overall} & \textbf{0.2840} & \underline{0.2831} & 0.2237 & \textbf{7.5891} & \underline{7.2862} & 6.6972  & \textbf{0.2740} & \underline{0.2688} & 0.2489\\
\bottomrule
\end{tabular}}
\caption{Comparison of IPGO+'s performance on \textbf{alignment, aesthetics and human preferences} with TextCraftor and DRaFT-1 across four datasets, COCO Person, COCO Mixed, DiffusionDB and PickaPic, for batch sizes $B=4$ and $B=10$. Boldface type indicates the highest scores, underline the second highest across all models for each of the three rewards.}
\vspace{-3mm}
\label{tab:batch_training}
\end{table}

\vspace{-5mm}
\subsection{Further Experiments and Ablation Studies}
\label{subsec:Further_Experiments}
\vspace{-2mm}
Additional experiments and ablation studies are provided in Appendices \ref{appx:Generalize_IPGO+_Tokens}, \ref{appx:adaptability_IPGO} and \ref{appx:abalation}. %Appendix \ref{appx:Generalize_IPGO+_Tokens} shows that the IPGO+ pre- and suffix tokens trained on the COCO Mixed data (which does not contain animals) effectively transfer a vivid style from the human preference training, and a learned water-painting style from the aesthetics training. 
\textbf{(1)} Appendix \ref{appx:Generalize_IPGO+_Tokens} shows that the IPGO+ pre- and suffix embeddings trained on the COCO Mixed data (without animals) effectively transfer styles from the reward models to unseen animal prompts. It also shows that convex combinations of IPGO+ embeddings trained on different rewards generate diverse images that smoothly mix different styles. \textbf{(2)} Appendix \ref{appx:adaptability_IPGO} demonstrates IPGO’s compatibility with SD1.5, SDXL, and SD3, improving human preference scores across all. \textbf{(3)} Appendix \ref{appx:abalation} provides a series of experiments motivating the setting of the hyperparameters $m_{\text{pre}}$ and $m_{\text{suff}}$ (the size of the orthonormal bases of the prefix/suffix $E_{\text{pre}}$/$E_{\text{suff}}$), the prefix/suffix lengths $N_{\text{pre}}$ and $N_{\text{suff}}$, activation of the \textit{Range} ($R$), \textit{Orthogonality} ($O$) and \textit{Conformity} ($C$) constraints, and the attention mechanism in IPGO+ for prompt-batch training. 
\vspace{-2mm}

\section{Conclusion}
\label{sec:conclusion}
\vspace{-2mm}
We propose IPGO(+), a parameter-efficient, gradient-based prompt-level optimization framework that improves diffusion models for better alignment with prompt semantics, aesthetics, and human preferences. IPGO explores, not alters, the prompt embedding space via learnable prefix and suffix embeddings, guided by reward gradients and constrained by range, orthonormality, and conformity. IPGO+ extends this with a parameter-free cross attention strengthening interactions among inserted prefix/suffix and the original prompt embeddings. Extensive experiments over seven benchmarks across three tasks, three datasets, and various diffusion model backbones (e.g. SDXL, SD3) demonstrate IPGO(+)'s strong performance gains on both prompt-wise training at inference and prompt-batch training, generalization to unseen prompts, and effectiveness of the IPGO(+) framework. We leave interpreting the optimized pre- and suffix embeddings as a topic for future research. 

% \begin{ack}
% Use unnumbered first level headings for the acknowledgments. All acknowledgments
% go at the end of the paper before the list of references. Moreover, you are required to declare
% funding (financial activities supporting the submitted work) and competing interests (related financial activities outside the submitted work).
% More information about this disclosure can be found at: \url{https://neurips.cc/Conferences/2025/PaperInformation/FundingDisclosure}.

% Do {\bf not} include this section in the anonymized submission, only in the final paper. You can use the \texttt{ack} environment provided in the style file to automatically hide this section in the anonymized submission.
% \end{ack}

\bibliography{references}
\bibliographystyle{plainnat}

%%%%%%%%%%%%%%%%%%%%%%%%%%%%%%%%%%%%%%%%%%%%%%%%%%%%%%%%%%%%
\newpage
\appendix
\section{Intuition on Optimization of Rotation Matrices}
\label{appx_sec: rotation}
\begin{figure}[!htb]
    \centering
    \includegraphics[width=0.9\linewidth]{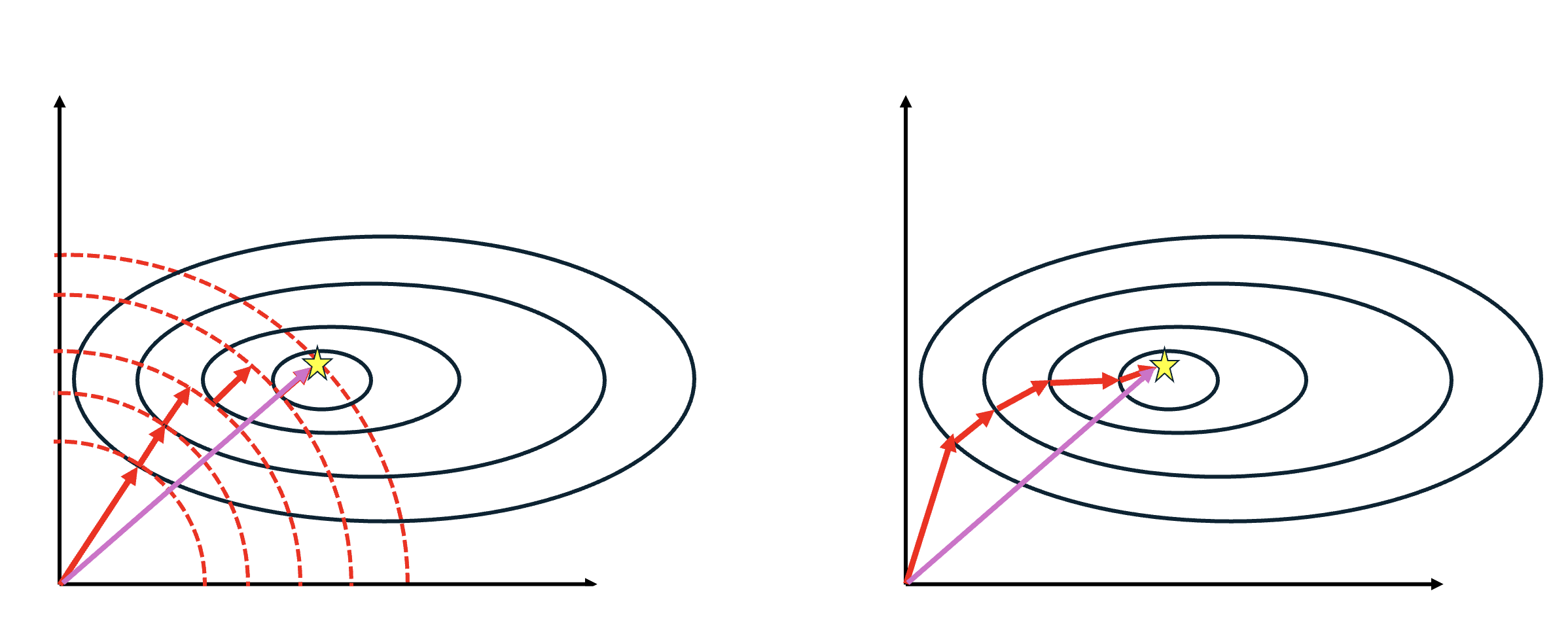}
    \caption{The figure illustrates the use of rotation in the optimization and compares the optimizations with rotation (left) and without rotation (right). The yellow star is the optimum point. The red dashed lines on the left are the circles with radius of the length of the current $x_t$. The purple line is the shortest distance between the initial point, the origin, and the optimum point. With rotations, the updates are calculated such that the total path is the shortest.}
    \label{fig:rotation_intuition}
\end{figure}

We use a toy example here to illustrate how the rotations help accelerate the optimization process. Suppose have a minimization problem 
\begin{equation}
    \text{argmin}_x f(x), \quad x\in\mathbb{R}^2.
\end{equation}
We assume this problem only has one global minimum $x^*$, which therefore lies in the subspace spanned by itself. Now we parameterize $x=R_{e,\theta}y$, with $y\in\mathbb{R}^2$ and $R_{e, \theta}$ the elementary rotation matrix in equation \ref{eqn:rotation_element}. Let us update parameters step by step. We initialize $x_0$ by $\theta_0=0$ and $y_0=0$ at the origin. We first move $x_0$ along the gradient of $x_0$ with a suitable step size to $x_1$, with $\theta_0$ unchanged. Assume we are currently at $x_t=R_{e,\theta_t}y_t$. We update $\theta_t$ by solving $\theta_{t+1}$ from:
\begin{equation}
    \nabla_xf(x_t)^T \frac{dR}{d\theta}\bigg|_{\theta_{t+1}}y_t=0.
\end{equation}
Note $\frac{dR}{d\theta}\big|_{\theta_{t+1}}=R_{e,\frac{\pi}{2}}R_{e,\theta_{t+1}}$. Therefore, graphically, the optimal $\theta_{t+1}$ is the one that rotates $y_t$, with the origin as the rotation axis, to a point such that the vector pointing to it is parallel to the gradient at that point. In other words, this is the point where the circle with radius $\|x_t\|$ is tangent to the contour of $f$. Then we use line search to find the suitable step size along the corrected gradient and move to the new point $x_{t+1}$. By this construction, the total path length to move from the origin $x_0$ to the optimal point $x_*$ is always the distance $\|x_*\|_2$, which is the shortest path between our initial point and the end point, therefore \textbf{optimal} among all possible pathways between the initial point to the optimum point. With this intuition, we add rotations to high dimension as well. Figure \ref{fig:rotation_intuition} visualizes the argument. The left panel shows a possible optimization path with rotation, which makes the total path length be exactly equal to the shortest path (the purple line) since the updated points are selected at the tangent point between the circles (red and dashed) and the contour. However, the regular gradient descent could easily take a longer path, as shown on the right panel.

\section{Qualitative Comparisons between IPGO(+) and Baselines}
Table \ref{tab:qualitative_compare} qualitatively compares our IPGO(+) algorithm with the benchmarking algorithms outlined in Section \ref{subsec:benchmarks}. As can be seen from the table, IPGO(+) is the only method that leverages reward gradient computation, supports prompt modification and batch training, and maintains low computational search costs.
\label{appx:qual_compare}
\begin{table}[!htb]
    \centering
    \resizebox{\textwidth}{!}{\begin{tabular}{l|ccccl}
    \toprule
        \textit{Algorithms} &  \textit{Reward Gradient} & \textit{Prompt Modification} & \textit{Batch Training} & \textit{Search Space Cost}\\
        \hline
        Promptist & \xmark  & \cmark & \cmark & High (SFT required) \\
        DPO-Diff & \cmark  & \cmark & \xmark & High (External LLM required)\\
        TextCraftor & \cmark  & \cmark & \cmark & High (Text Encoder Fine-Tuning)\\
        DRaFT & \cmark  & \xmark & \cmark & Low\\
        DDPO & \xmark  & \xmark & \cmark & High (Many samples required)\\
        ChatGPT-4o & \xmark  & \cmark & \xmark & Low\\
        \hline
        IPGO (ours) & \cmark  & \cmark & \cmark & Low \\
        \bottomrule
    \end{tabular}}
    \caption{A qualitative comparison between IPGO(+) and all benchmarks, focusing on four key aspects: the ability to compute reward gradients, support for prompt modification, compatibility with batch training, and the computational cost of searching the prompt space.}
    \label{tab:qualitative_compare}
\end{table}

\section{Implementation Details}
\label{appx:hyperparameters}
This section provides details on IPGO(+)'s implementation and training. 

\paragraph{Image Generation} A DDIM Scheduler with a guidance weight of 7.5 is employed and the generated images have a resolution of $512\times512$ pixels. During optimization, we truncate the backpropagation at the second last sampling step. For prompt-wise training, we set sampling step as 50. For prompt-batch training, we set sampling step as 30. We found similar performances with other sampling strategies, such as PNDM and LMSD. 

\paragraph{Optimization} We train IPGO using Adam \cite{kingma2014adam} optimizer without a weight decay. For prompt-wise training, we start with a learning rate of $1e-3$ and reduce it by a factor of $0.9$ every 10 epochs, continuing this schedule for a total of 50 epochs. We truncate gradients at the 2nd-to-last step with checkpointing. For prompt-batch training, we use a cosine scheduling between $1e-4$ and $1e-5$. We apply gradient clipping across all our experiments, selecting a gradient clipping norm of $c=1.0$. We truncate gradients at the 10th-to-last step with checkpointing.

\paragraph{Hyperparameters} We set the hyperparameters of IPGO and DRaFT-1 to ensure that the total number of trainable parameters is comparable. IPGO(+) includes the hyperparameters $N$, $M_{pre}$ and $M_{suf}$. TextCraftor uses its default configuration. DRaFT-1 has a low rank for the LoRA parameters in the UNet. DDPO also uses the default configurations. Detailed hyperparameter settings for IPGO, TextCraftor, DRaFT-1 and DDPO are provided in Table \ref{tab:hyperparameters}. 
\begin{table}[!h]
\centering
\begin{tabular}{l|cc}
\toprule
\textit{Methods}       & \textit{Hyperparameter}                & \textit{Value}                 \\ \midrule
IPGO         & $m_{\text{pre}},m_{\text{suff}}$                             & 300                    \\
             & $N_{\text{pre}}$                        & 10                    \\
             & $N_{\text{suff}}$                        & 10                    \\
\hline
Total \#parameters &                               & 0.47M                \\ \hline
\midrule
TextCraftor & \multicolumn{2}{l}{Default Configuration} \\ \hline
Total \#parameters & & 123M \\ 
\hline
\midrule
DRaFT-1      & LoRA rank                     & 3                     \\
\hline
Total \#parameters &                               & 0.60M                \\ \hline
\midrule
DDPO         & \multicolumn{2}{l}{Default DDPOTrainer Configuration} \\ 
\hline
Total \#parameters & & 0.79M \\ 

\bottomrule
\end{tabular}
\caption{Hyperparameter settings for IPGO, TextCraftor, DRaFT-1 and DDPO}
\label{tab:hyperparameters}
\end{table}

\paragraph{IPGO(+)'s Constraints} IPGO(+) has three constraints: Orthogonality, Value and Conformity constraints. We enforce the orthogonality constraint with {\fontfamily{qcr}\selectfont
orthogonal()} module in Pytorch. For the value constraint, we clamp the parameters to satisfy the constraint after each update. Finally, for simplicity, we use a soft conformity constraint in the optimization instead of a hard one. We add a conformity penalty to the objective, the negative image reward. Define the conformity penalty by
\begin{equation}
\label{eqn:conform_penalty}
% p_{conf}=\left\|\frac{1}{J}\sum_{j=1}^J T_j - \frac{1}{K}\sum_{k=1}^K f(C)_k\right\|^2_2,
p_{\text{conf}} = \|\text{Mean}(\mathcal{E}(V_{\text{pre}},p,V_{\text{suff}};~\Omega_{\text{IPGO}}))-\text{Mean}(\mathcal{T}(p))\|_2^2
\end{equation}
from equation \ref{eqn:sum constraint}. Then the optimization objective becomes:
\begin{equation}
\label{eqn:objective}
\mathcal{L} = -\mathcal{S}(x_0,p) + \gamma p_{\text{conf}},    
\end{equation}
where $\mathcal{S}(x_0,p)$ is one of the Aesthetic, CLIP and HPSv2 reward scores as a function of the image $x_0$ and prompt $p$ (or their weighted combination), and $\gamma$ is the conformity coefficient. Our experiments set $\gamma=1e-3$.

\paragraph{The Outline of Our IPGO Algorithm} ~

\begin{algorithm}[!h]
   \caption{IPGO(+)}
   \label{alg:IPGO_algo}
\begin{algorithmic}
   \State {\bfseries Input:} Raw prompt $p$, prefix/suffix generator $G_{\text{pre}}(\Omega_{\text{IPGO}})$/$G_{\text{suff}}(\Omega_{\text{IPGO}})$ controlled by $\Omega_{\text{IPGO}}$, text encoder $\mathcal{T}(p)$, diffusion model $x\sim q_{\text{image}}(\cdot)$, $z_T$ the initial latent noise, image reward model $S(x,p)$, conformity penalty coefficient $\gamma$, learning rate $\eta$, number of epochs $Epochs$
   \State {\bfseries Output:} Optimal prefix/suffix generators $G^*_{\text{pre}}$/$G^*_{\text{suff}}$.
   \For{$i=0$ {\bfseries to} $Epochs$}
   \State Original prompt embedding: $V_0=\mathcal{T}(p)$.
   \State Prefix: $V_{\text{pre}}=G_{\text{pre}}(\Omega_{\text{IPGO}})$
   \State Suffix: $V_{\text{suff}}=G_{\text{suff}}(\Omega_{\text{IPGO}})$
   \If{IPGO+}
   \State $V_{\text{pre}}\leftarrow V_{\text{pre}}+\text{Attention}(V_{\text{pre}},\mathcal{T}(p),\mathcal{T}(p))$
   \State $V_{\text{suff}}\leftarrow V_{\text{suff}}+\text{Attention}(V_{\text{suff}},\mathcal{T}(p),\mathcal{T}(p))$
   \EndIf
   \State Insert prefix and suffix: $\mathcal{E}(V_{\text{pre}},p,V_{\text{suff}};~\Omega_{\text{IPGO}})=V_{\text{pre}}\oplus V_0\oplus V_{\text{suff}}$.
   \State Sample image: $x_0\sim q_{\text{image}}(x_0|\mathcal{E}(V_{\text{pre}},p,V_{\text{suff}};~\Omega_{\text{IPGO}}),z_T)$.
   \State Compute reward: $r=\mathcal{S}(x_0,p)$.
   \State Compute objective: $\mathcal{L}=-r+\gamma p_{\text{conf}}$.
   \State Compute gradient: $g = \nabla_{\Omega_{\text{IPGO}}}\mathcal{L}$.
   \State Update prefix and suffix: $\Omega_{\text{IPGO}} \leftarrow \Omega_{\text{IPGO}}-\eta g$.
   \State Enforce Orthogonality and Value constraints.
   \EndFor
   \State Return $G^*_{\text{pre}}(\Omega_{\text{IPGO}})$ and $G^*_{\text{suff}}(\Omega_{\text{IPGO}})$.
\end{algorithmic}
\end{algorithm}

\section{Adaptability of IPGO to other Diffusion Models}
\label{appx:adaptability_IPGO}
We next illustrate that IPGO is not dependent on a single diffusion model, but can be used with different diffusion models. In the experiments reported heretofore we choose to implement IPGO(+) with SD-v1.5 because this diffusion model performs well in real-world human evaluations, has a comparatively smaller number of parameters and more attractive  computation characteristics as compared to newer diffusion models. Nonetheless, IPGO(+) can be used to improve image quality at inference for a wider range of diffusion models. Here, we illustrate IPGO for newer versions of Stable Diffusion, SDv1.5, SDXL, and SD3, for HPSv2 human preference rewards on 100 randomly selected prompts from the DiffusionDB data. Table~\ref{tab:IPGO_other_SDs} shows the results. 

Table~\ref{tab:IPGO_other_SDs} shows that IPGO improves human preference scores for each of the three diffusion models, SD1.5, SD3 and SDXL, with the largest improvement being for SDXL, 4.9\%.

\begin{table}[!htb]
\centering
\begin{tabular}{lccc}
\toprule
 Diffusion model      & SDXL   & SD3  & SDv1.5  \\
\midrule
Original    & 0.2523 & 0.2625 & 0.2621\\
w/ IPGO &  0.2646 & 0.2710 & 0.2729\\
\hline 
Improvement & 0.0123 & 0.0085 & 0.0108\\
\bottomrule
\end{tabular}
\caption{IPGO on other Diffusion Models on the HPSv2 reward with respect to 100 randomly selected prompts from our DiffusionDB prompt dataset.}
\label{tab:IPGO_other_SDs}
\end{table}

\section{Generalizability of Learned IPGO+ Prefix and Suffix}
\label{appx:Generalize_IPGO+_Tokens}
\begin{figure*}[!h]
    \centering
    \includegraphics[width=1.0\linewidth]{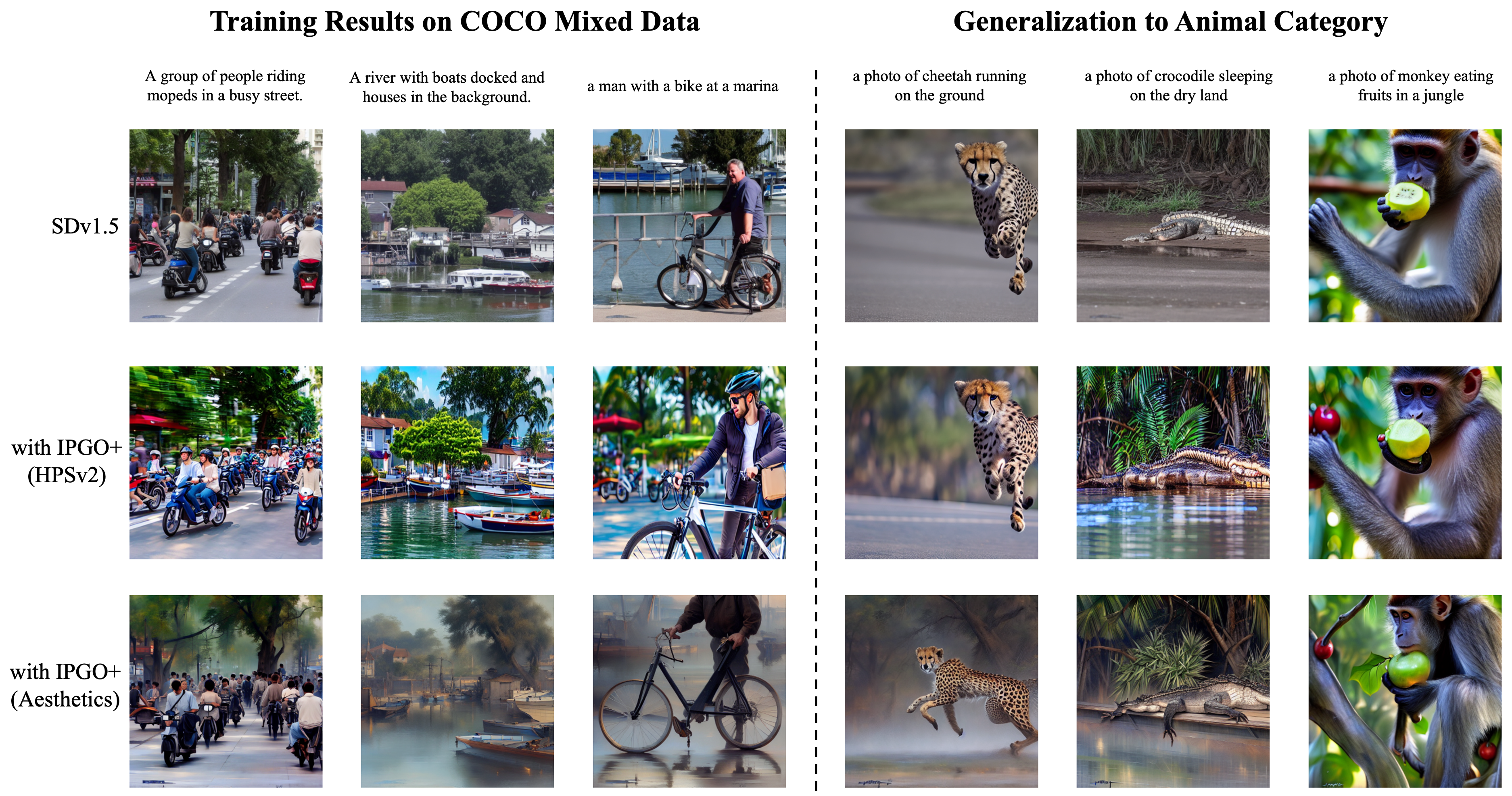}
    \vspace{-4mm}
    \caption{Example images generated by prompt-batch training with IPGO+ on the HPSv2 and aesthetic rewards. The left three columns are the resulting images after training on the COCO Mixed data; the right three columns show images resulting from inserting the trained prefix and suffix in the unseen animal category prompts.}
    \label{fig:batch_aes_images}
\end{figure*}

The first row in Figure \ref{fig:batch_aes_images} presents images generated from raw prompts, while the second and third rows display images generated from prompts with IPGO+ prefix and suffix optimized for human preference and aesthetics rewards respectively. The left three columns show images generated from prompts in the COCO Mixed training data which do not involve any animals, whereas the right columns show images generated from three prompts in the COCO animals category. Pre- and suffix embeddings trained on the mixed data were inserted into the prompt embeddings for the animals category before the images are generated. The figure illustrates that a learned vivid style from the human preference training, and a learned water-painting style from the aesthetics training are effectively transferred, via the IPGO+ prefix and suffix, to the unseen animal prompts. These results illustrate good generalizability of IPGO+ to unseen prompts. 

\begin{figure}[!h]
    \centering
    \includegraphics[width=0.8\linewidth]{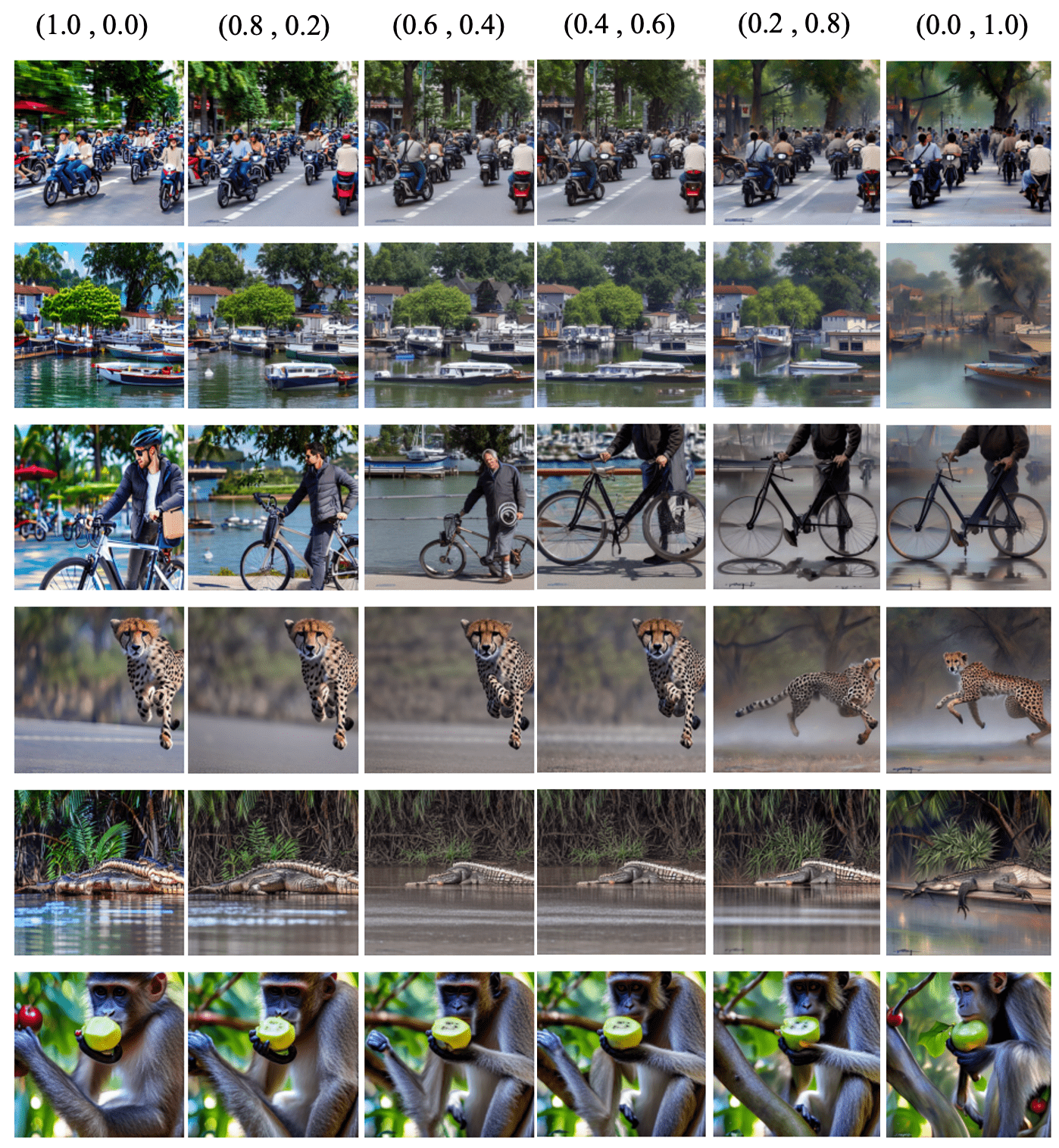}
    \caption{The rows show six sets of example images (see the columns of Figure~\ref{fig:batch_aes_images} generated from prompts that use convex combinations of the prefixes/suffixes from human preference and aesthetics rewards. The top row indicates the combination weights, in the format of (human preference weight, aesthetics weight). We find the combined style smoothly changes from purely human-preference-styled to purely aesthetics-styled.}
    \label{fig:convex_combination}
\end{figure}

Figure \ref{fig:convex_combination} shows images with mixed styles, resulting from  convex combinations of the trained prefixes and suffixes trained on human preference and aesthetics, respectively. The style of the image shifts smoothly from the style implied by the human preference reward to that by the aesthetics reward, \textit{without} distorting the semantics. Therefore, the figure illustrates the feasibility of post-processing image styles by convex mixing of the prefix and suffix learned from various reward models.

\section{Ablation Studies}
\label{appx:abalation}
We provide an in-depth analysis of our design choices, including low rank representation and rotation, as well as hyperparameters: $M$, the number of vectors in the orthonormal bases of the prefix ($E_{\text{pre}}$) and suffix ($E_{\text{suff}}$); the lengths of the prefix ($N_{\text{pre}}$) and suffix ($N_{\text{suff}}$) respectively; activation of the three constraints: \textit{Range} ($R$), \textit{Orthogonality} ($O$) and \textit{Conformity} ($C$) constraints; and finally a comparison between IPGO and IPGO+ in the prompt-wise and prompt-batch training scenarios.

\paragraph{Low Rank Representation and Rotation.} Embedding learning may offer an alternative to prefix and suffix learning, which we investigate on 60 prompts randomly selected from the COCO dataset. Evaluation is conducted using CLIP rewards averaged across all prompts. In addition, we examine the impact of rotation on these 60 prompts by comparing IPGOs performance with and without the rotation. The experimental results show that IPGO with rotation achieves a higher average CLIP score of 0.297 than the 0.295 score obtained without rotation, and both significantly outperform the simple embedding scheme which scores 0.259.

\paragraph{Constraints.} To investigate the impact of the $R$, $O$ and $C$ constraints on the optimization, we conduct a series of ablation studies. Using the same set of 60 randomly selected prompts as before,  we evaluate the average CLIP reward and progressively test combinations of constraints. First, the \textit{Range} constraint significantly enhances optimization, as evidenced by the substantial improvements in CLIP reward when comparing results without the constraint (0.277) to those with the \textit{Range} constraint alone (0.310). Second, while adding the single \textit{Conformity} constraint to the \textit{Range} constraint degrades performance (0.301 v.s. 0.310), additionally imposing the \textit{Orthogonality} constraint substantially benefits  overall optimization (0.322 v.s. 0.310). Therefore, incorporating all three constraints into the search space optimizes the efficiency of our algorithm, by introducing nonlinearity, enlarging the search space, helping to ensure a stable solution, reducing the likelihood of exploring extreme text embeddings, and increasing the CLIP reward. Consequently, we employ all three constraints after rotation in our experiments.

\paragraph{Varying $m=m_{\text{pre}}=m_{\text{suff}}$.} We investigate the effect of $m_{*}$, the size of the learnable prefix and suffix prompt embeddings, on CLIP reward optimization. Using 30 randomly selected prompts from the COCO dataset, we conduct ablation studies with different values of $m$. We find that the optimal performance occurs at $m=300$ $(m=50: 0.2791, m=100: 0.2820, m=200: 0.2863, m=300: 0.2872, m=400: 0.2845)$
%(see Figure \ref{fig:ablation_N} in Appendix D). 
Notably, increasing the number of inserted embeddings does not necessarily lead to proportional improvements in performance.

\begin{figure}[!h]
    \centering
    \includegraphics[width=0.35\textwidth]{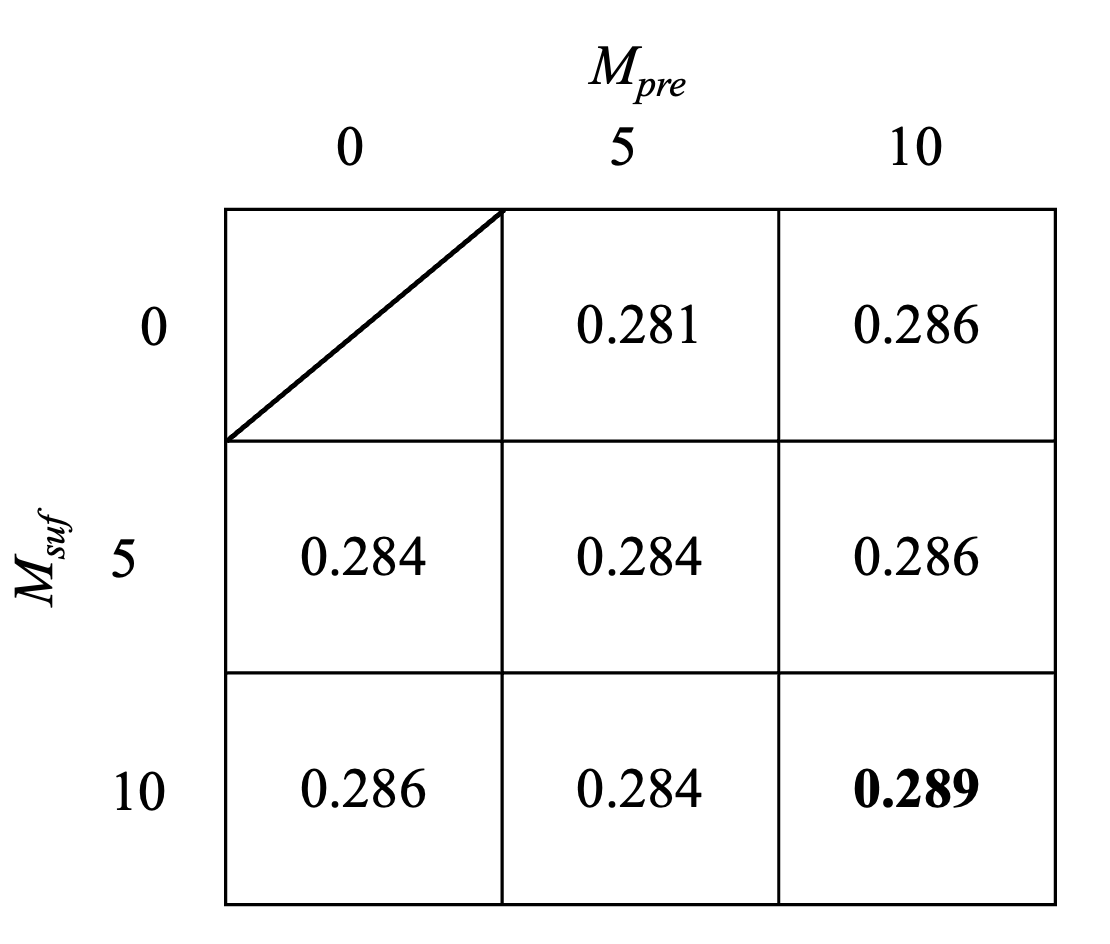}
    \caption{Experiments on $M_{pre}$ and $M_{suf}$. The table shows that a balanced prefix and suffix length at 10 yields the best performance.}
    \label{fig:ablation_Mpre_Msuf_table}
\end{figure}

\paragraph{$N_{\text{pre}}$ and $N_{\text{suff}}$ based on Raw Prompt Length.} Next, we investigate the optimal prefix and suffix lengths. We test all combinations of prefix and suffix lengths of 0, 5, or 10 embeddings, excluding the $(0,0)$ combination. Figure \ref{fig:ablation_Mpre_Msuf_table} presents a table that contains average CLIP scores of all possible combinations of $M_{pre}$ and $M_{suf} \in \{0,5,10\} $ in our experiment. We observe that a balance of the prefix and suffix lengths tends to give a better performance. For combinations that have 5 (or 10) as the maximum length, $(5,5)$ (or $(10,10)$) always yields the best performance, with the latter being the highest. 

Next, we test the relationship between the length of the raw prompt and the lengths of prefix and suffix. We choose 30 prompts among which the first 10 prompts are simple prompts, such as ``Man", ``Woman" and ``Student", the second 10 prompts are medium-complexity prompts, selected from the COCO dataset, and the last 10 prompts are even more complex versions of the second 10 prompts by inquiring ChatGPT-4o with ``Could you make the following 10 prompts more complex:." For example, the complex version of ``A person walking in the rain while holding an umbrella." is ``A middle-aged person in a long, tattered trench coat walks down a cobblestone street, their brightly colored umbrella catching the dim glow of streetlights as rain cascades around them." We make sure that the complex sentences do not exceed the limit of 77 tokens. We optimize each prompt with $M_{pre}=M_{suf}=M \in \{2,10,15\}$ with respect to the CLIP reward. We use the distribution of the number of prompts that respectively have $M=2,10,15$ as their best prefix and suffix lengths in each prompt group as the evaluation metric. For simplicity, we denote this distribution as $D_M(2,10,15)$.

We do not observe a significant correlation between the prompt length and the prefix and suffix lengths. The simple prompt group has $D_M(2,10,15)=(30\%,50\%,20\%)$; the medium-complex prompt group has $D_M(2,10,15)=(50\%,20\%,30\%)$; and the complex prompt group has $D_M(2,10,15)=(20\%,40\%,40\%)$. We observe a weak positive relationship of $M$ with the raw prompt length at best. We find no correlation between the raw prompt length and the optimal prefix-suffix length. However, we recommend using fewer inserted embeddings for very short prompts to avoid over-parameterization (an illustration follows).

%Using the same 30 randomly selected prompts and evaluating with CLIP reward, we find that the $(10,10)$ combination yields the highest score of 0.289, while $(0,10)$ and $(10,0)$, corresponding to a setup with \textit{only the suffix, respectively only the prefix}, achieve lower scores of 0.286 each, which reveals that both the prefix and suffix contribute to the alignment (see Figure~\ref{fig:ablation_Mpre_Msuf_table} in Appendix D). %These results suggest that performance is improved when the lengths of prefix and suffix are balanced and equal. 
%In addition, we explore the relationship between the raw prompt length and the prefix-suffix length, and %. We use 30 prompts, comprising 10 prompts with only one or two tokens, 10 prompts with approximately 10-15 tokens, and 10 prompts with around 25 tokens. We 

\paragraph{IPGO v.s. IPGO+} Finally, we compare IPGO and IPGO+ in the two training scenarios for human preference rewards, since IPGO(+) has the most consistent performance with this reward. For prompt-wise training, we randomly pick 30 prompts from each of the three datasets. IPGO on average achieves 0.2811, higher than IPGO+'s 0.2764. For prompt-batch training, we use the 60 prompts from the COCO-person dataset. We find that IPGO achieves 0.2785 (averaged across batch sizes $B=4,10$) after 1200 image generations, but IPGO+ achieves a much higher 0.2901. These results illustrate that IPGO is better at prompt-wise training, while IPGO+ is better and the cross attention is more effective at prompt-batch training.

%\section{Details of Ablation Studies}
%\paragraph{$M_{pre}$ and $M_{suf}$ based on Raw Prompt Length} 

\begin{figure}[!h]
    \centering
    \includegraphics[width=0.8\linewidth]{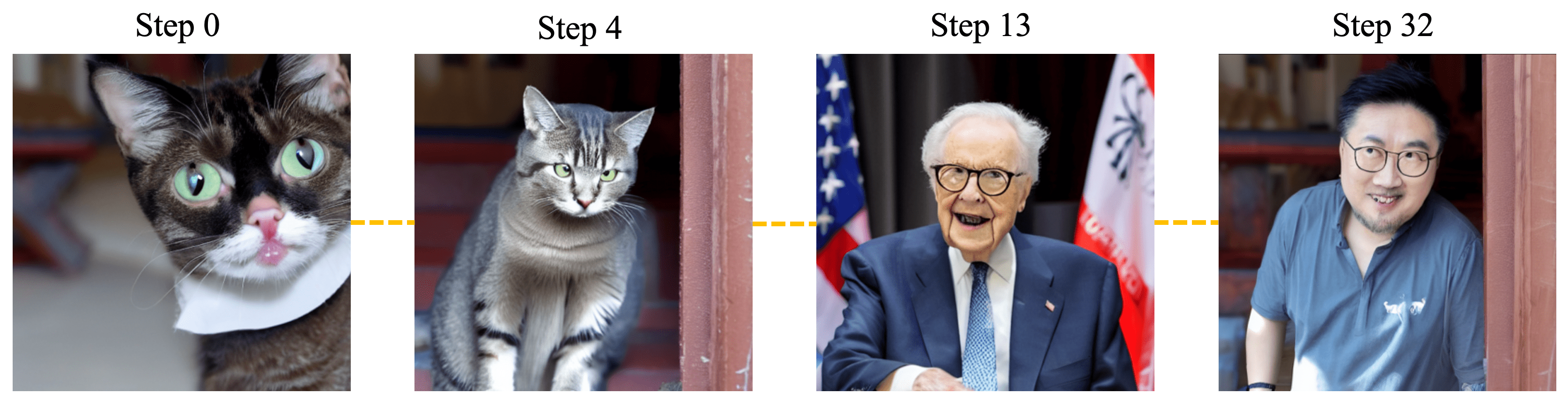}
    \caption{Images generated during IPGO's optimization on the prompt ``Cat" with $N_{\text{pre}}=N_{\text{suff}}=30$ for aesthetics. Because of  overparameterization the images that are produced show poor alignment with the \textit{original} prompt.}
    \label{fig:overparameterization}
\end{figure}

\paragraph{Overparameterization} IPGO faces the risk of overparameterization when the lengths of the prefix and suffix significantly exceed that of the raw prompt. For example, Figure \ref{fig:overparameterization} illustrates this issue with an extreme example, showcasing the evolution of images generated during the optimization of the simple prompt ``cat" with very long $N_{\text{pre}}=N_{\text{suff}}=30$ for aesthetics improvement. In the first several steps, IPGO produces images that display a cat, but at later steps, the object in the image changes to a person, and at even later steps the specific person also changes. Apparently, if the prefix and suffix are too long, then in spite of the conformity constraint, optimization of  the inserted embeddings overwhelms the semantic structure of the image and harms alignment of the image with the \textit{original} prompt. Therefore, we recommend using shorter prefix and suffix lengths for shorter prompts.

\section{Societal Impact}
\label{appx:impact_statement}
This paper presents work that contributes to the field of Text-to-Image generation models and their applications. In the Machine Learning community, the new method introduced by this paper can broaden the current horizon on fine-tuning diffusion models. In practice, our method can be applied to image related tasks such as automatic real-time image editing.

\newpage
\section{Additional Image Examples}
\label{appx:more_images}
\subsection{Additional IPGO Comparisons with DRaFT-1 and TextCraftor}
\label{appx:IPGO_extra_images}
\begin{figure}[!h]
    \centering
    \includegraphics[width=0.9\linewidth]{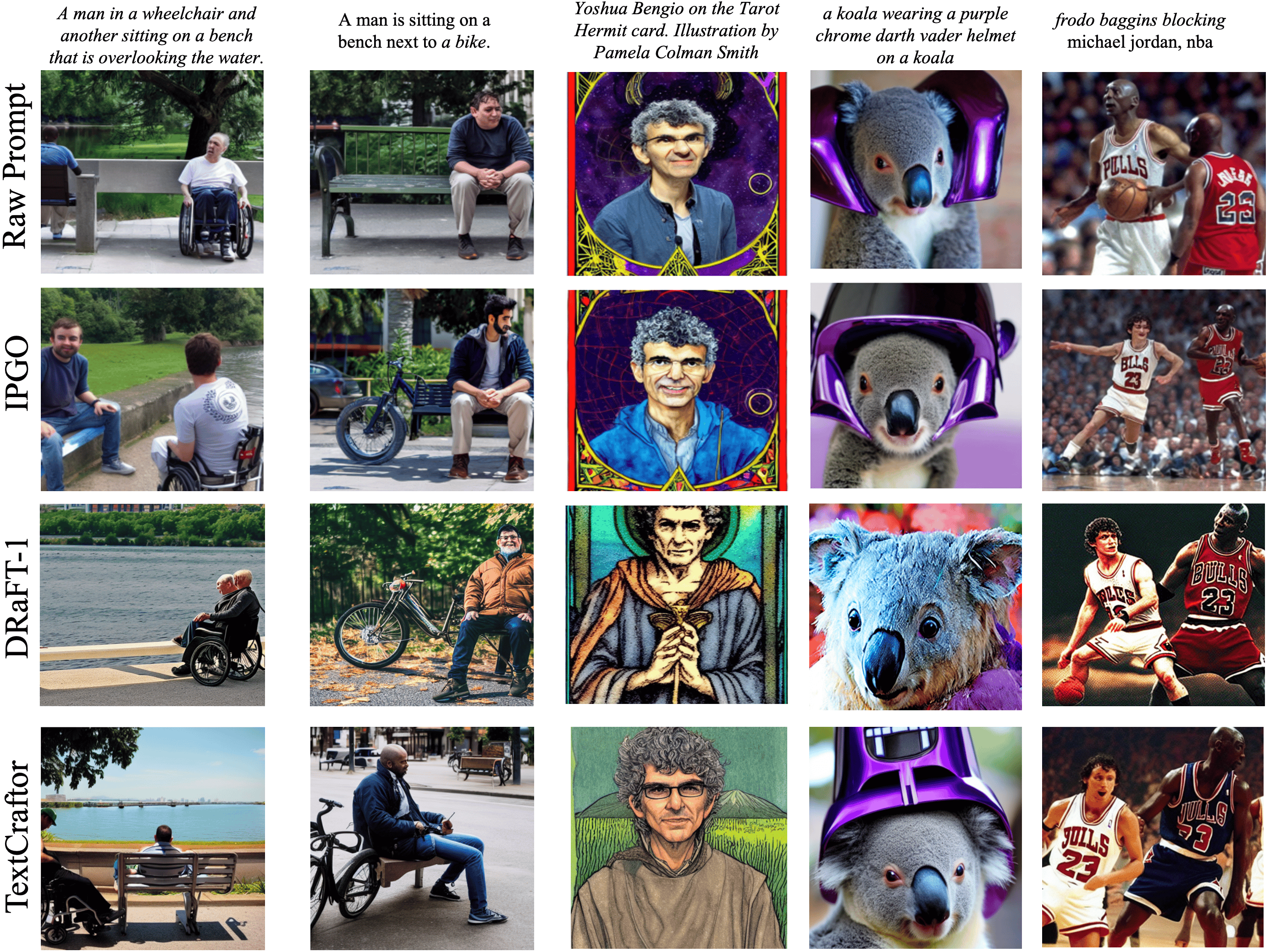}
\end{figure}
\begin{figure}[!h]
    \centering
    \includegraphics[width=0.9\linewidth]{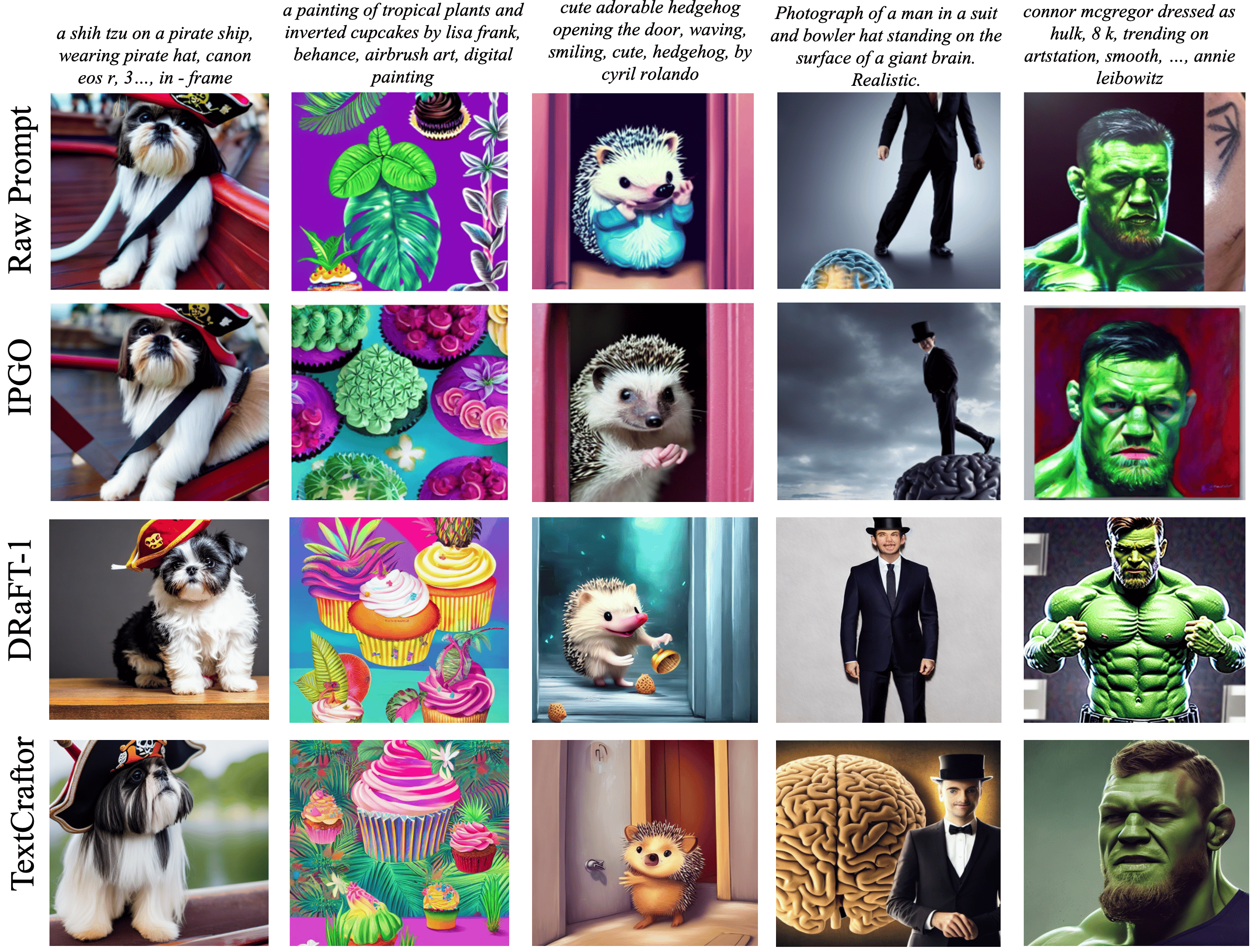}
\end{figure}

\newpage
\subsection{IPGO+ Comparisons with TextCraftor}
\label{appx:IPGO+_extra_images}
\begin{figure}[!h]
    \centering
    \includegraphics[width=1.0\linewidth]{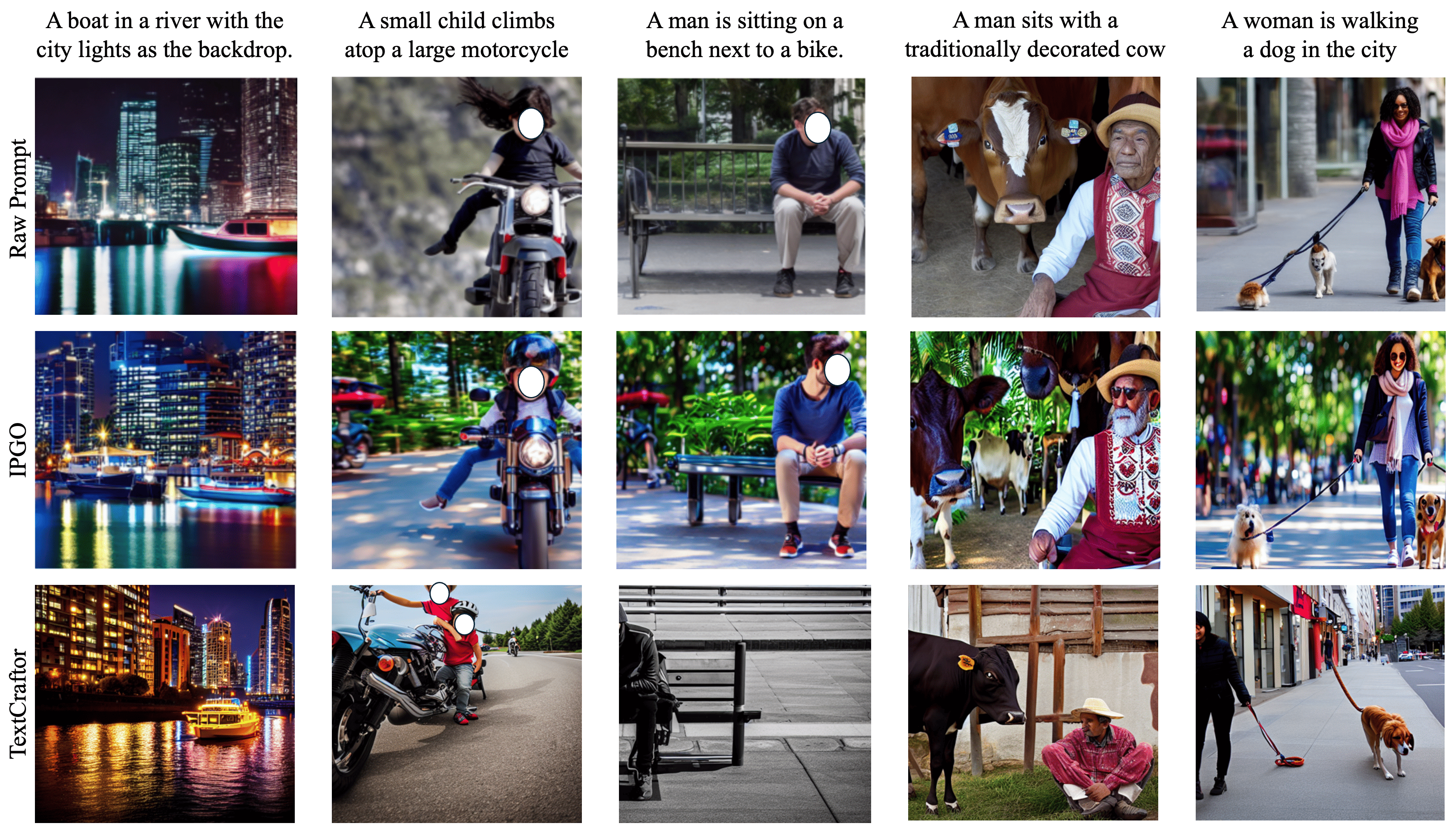}
    % \caption{Caption}
    % \label{fig:enter-label}
\end{figure}
\begin{figure}[!h]
    \centering
    \includegraphics[width=1.0\linewidth]{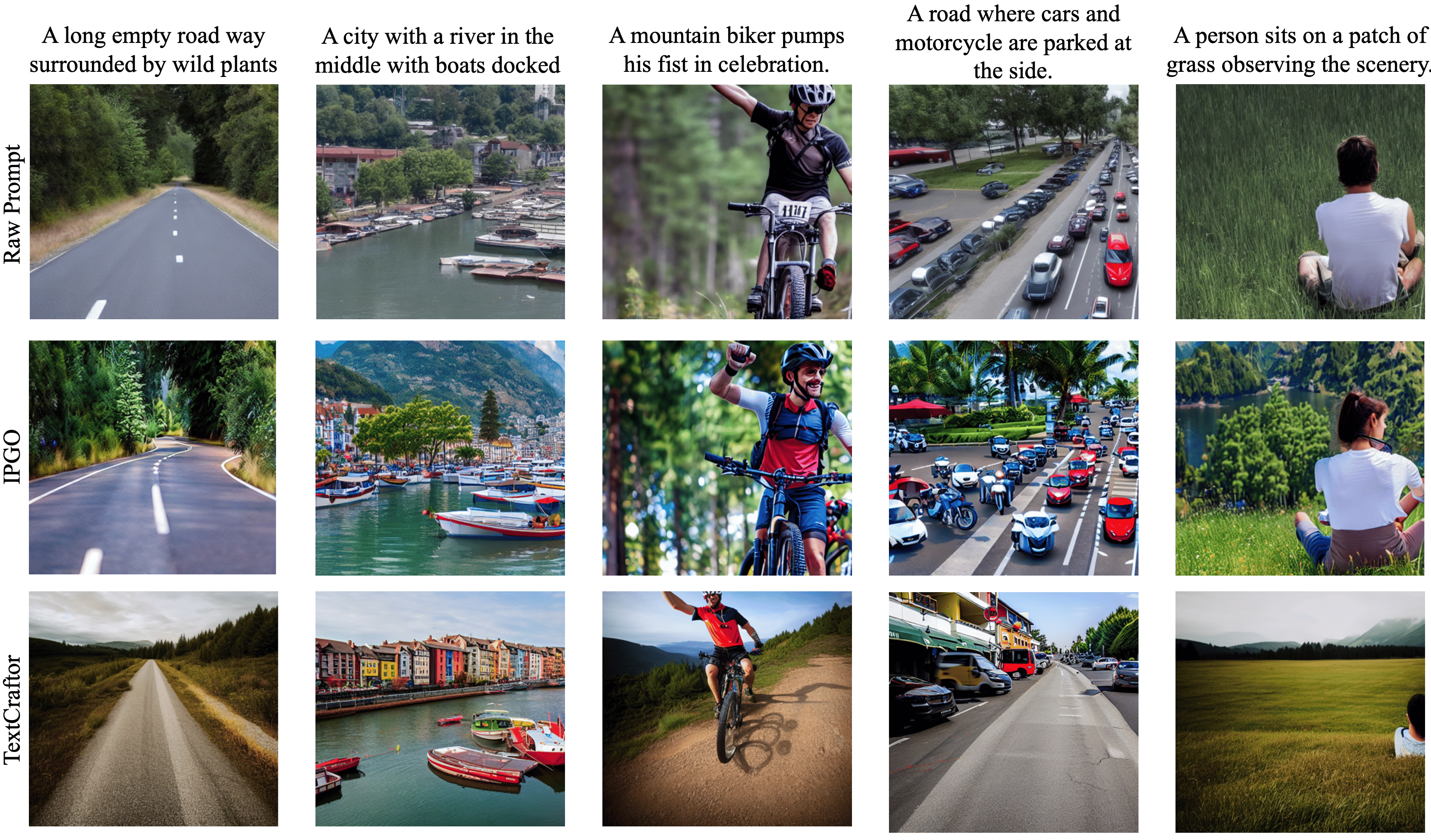}
    % \caption{Caption}
    % \label{fig:enter-label}
\end{figure}

\newpage
\subsection{Others}
\label{appx:IPGO_more_images_others}
\begin{figure}[!h]
    \centering
    \includegraphics[width=1.1\linewidth]{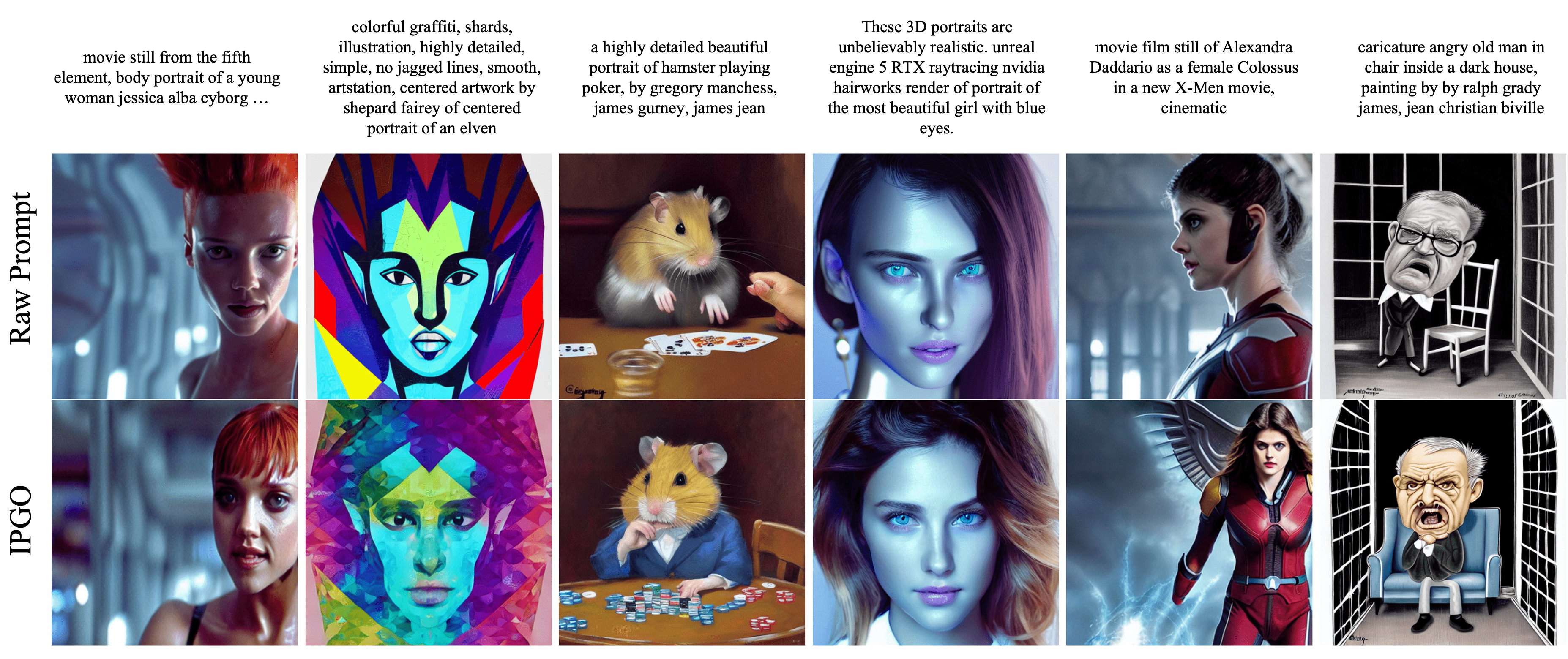}
\end{figure}
\begin{figure}[!h]
    \centering
    \includegraphics[width=1.1\linewidth]{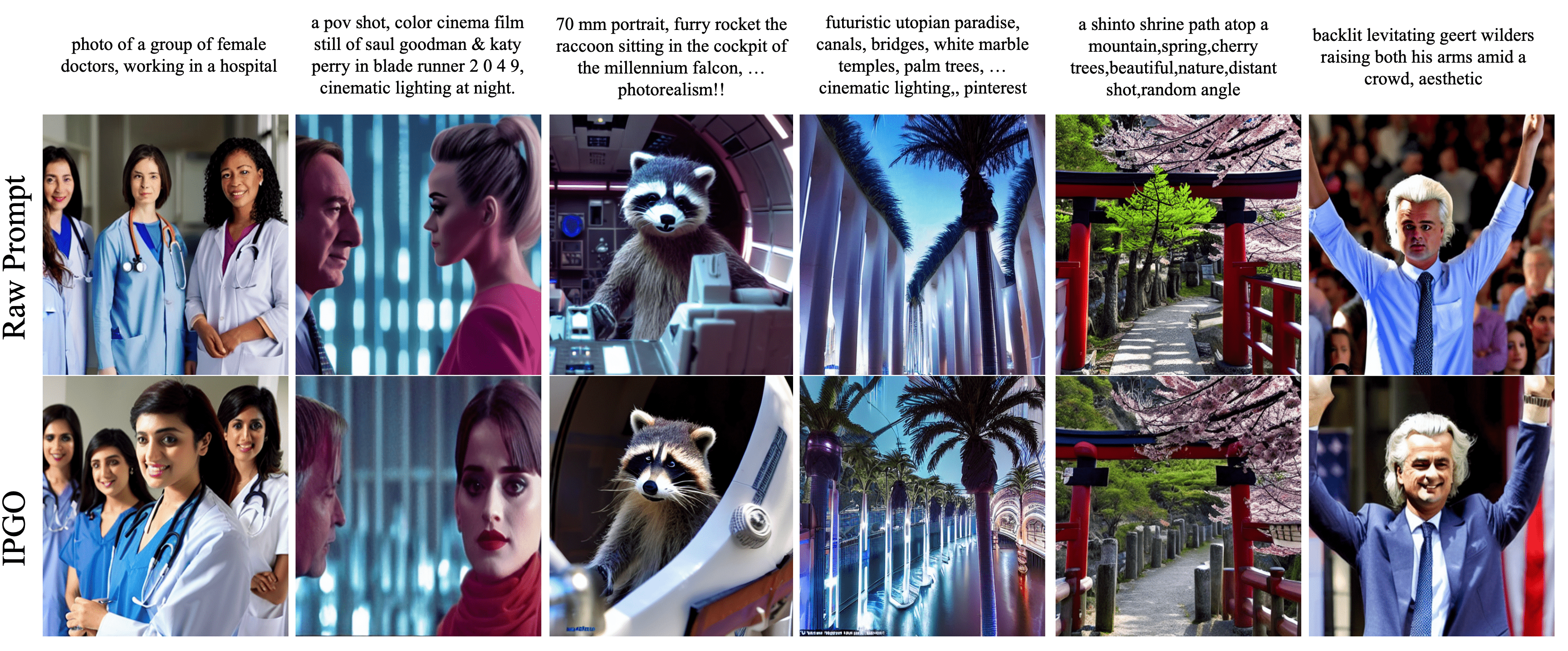}
\end{figure}
\begin{figure}[!h]
    \centering
    \includegraphics[width=1.1\linewidth]{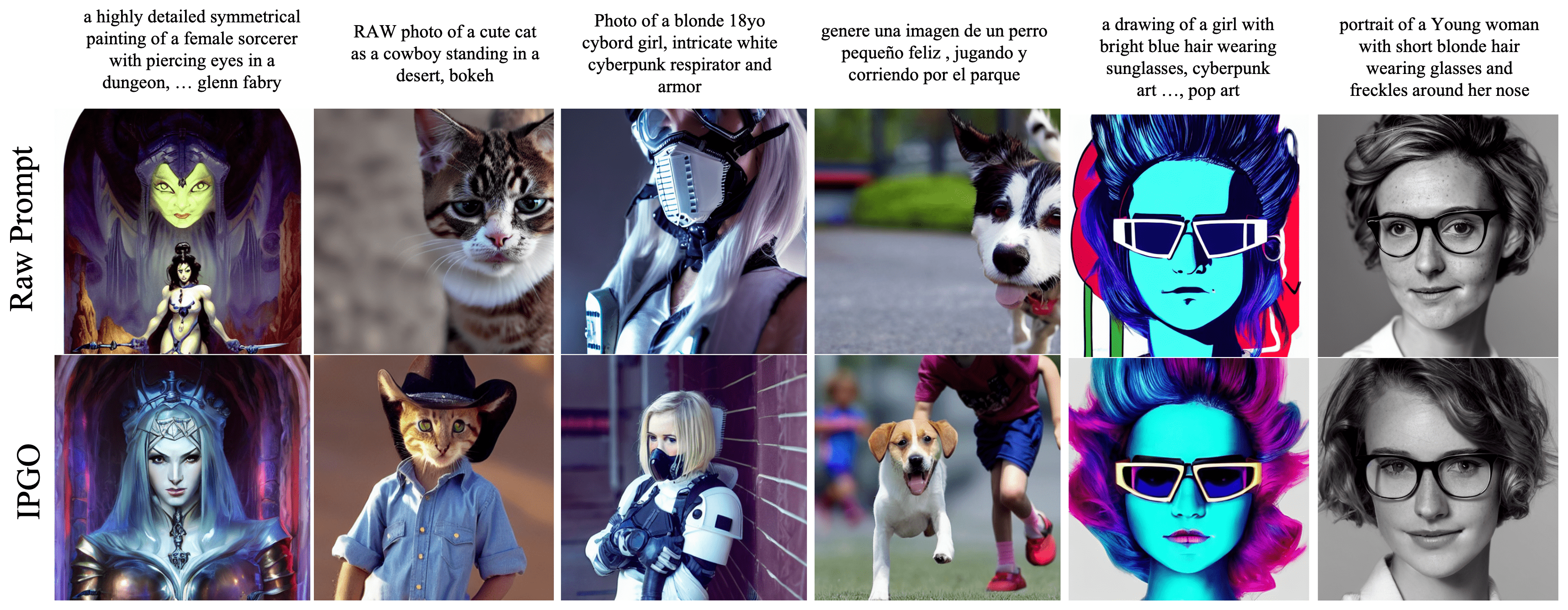}
\end{figure}

%%%%%%%%%%%%%%%%%%%%%%%%%%%%%%%%%%%%%%%%%%%%%%%%%%%%%%%%%%%%

\newpage
\section*{NeurIPS Paper Checklist}

%%% BEGIN INSTRUCTIONS %%%
The checklist is designed to encourage best practices for responsible machine learning research, addressing issues of reproducibility, transparency, research ethics, and societal impact. Do not remove the checklist: {\bf The papers not including the checklist will be desk rejected.} The checklist should follow the references and follow the (optional) supplemental material.  The checklist does NOT count towards the page
limit. 

Please read the checklist guidelines carefully for information on how to answer these questions. For each question in the checklist:
\begin{itemize}
    \item You should answer \answerYes{}, \answerNo{}, or \answerNA{}.
    \item \answerNA{} means either that the question is Not Applicable for that particular paper or the relevant information is Not Available.
    \item Please provide a short (1–2 sentence) justification right after your answer (even for NA). 
   % \item {\bf The papers not including the checklist will be desk rejected.}
\end{itemize}

{\bf The checklist answers are an integral part of your paper submission.} They are visible to the reviewers, area chairs, senior area chairs, and ethics reviewers. You will be asked to also include it (after eventual revisions) with the final version of your paper, and its final version will be published with the paper.

The reviewers of your paper will be asked to use the checklist as one of the factors in their evaluation. While "\answerYes{}" is generally preferable to "\answerNo{}", it is perfectly acceptable to answer "\answerNo{}" provided a proper justification is given (e.g., "error bars are not reported because it would be too computationally expensive" or "we were unable to find the license for the dataset we used"). In general, answering "\answerNo{}" or "\answerNA{}" is not grounds for rejection. While the questions are phrased in a binary way, we acknowledge that the true answer is often more nuanced, so please just use your best judgment and write a justification to elaborate. All supporting evidence can appear either in the main paper or the supplemental material, provided in appendix. If you answer \answerYes{} to a question, in the justification please point to the section(s) where related material for the question can be found.

IMPORTANT, please:
\begin{itemize}
    \item {\bf Delete this instruction block, but keep the section heading ``NeurIPS Paper Checklist"},
    \item  {\bf Keep the checklist subsection headings, questions/answers and guidelines below.}
    \item {\bf Do not modify the questions and only use the provided macros for your answers}.
\end{itemize}

%%% END INSTRUCTIONS %%%

\begin{enumerate}

\item {\bf Claims}
    \item[] Question: Do the main claims made in the abstract and introduction accurately reflect the paper's contributions and scope?
    \item[] Answer: \answerYes{} % Replace by \answerYes{}, \answerNo{}, or \answerNA{}.
    \item[] Justification: This paper introduces a new parameter-efficient framework for fine-tuning Text-to-Image diffusion models that can outperform SOTA methods of various categories. See Sec.\ref{sec:methods} and Sec.\ref{sec:experiments}.

\item {\bf Limitations}
    \item[] Question: Does the paper discuss the limitations of the work performed by the authors?
    \item[] Answer: \answerYes{} % Replace by \answerYes{}, \answerNo{}, or \answerNA{}.
    \item[] Justification: See Sec.\ref{sec:conclusion} and Appendix \ref{appx:abalation}.

\item {\bf Theory assumptions and proofs}
    \item[] Question: For each theoretical result, does the paper provide the full set of assumptions and a complete (and correct) proof?
    \item[] Answer: \answerNA{} % Replace by \answerYes{}, \answerNo{}, or \answerNA{}.
    \item[] Justification: This paper has no theoretical results.

    \item {\bf Experimental result reproducibility}
    \item[] Question: Does the paper fully disclose all the information needed to reproduce the main experimental results of the paper to the extent that it affects the main claims and/or conclusions of the paper (regardless of whether the code and data are provided or not)?
    \item[] Answer: \answerYes{} % Replace by \answerYes{}, \answerNo{}, or \answerNA{}.
    \item[] Justification: We include an anonymous gihub link that includes codes for reproducibility. For more details about the algorithm and training configurations, see Appendix \ref{appx:hyperparameters} and Appendix

\item {\bf Open access to data and code}
    \item[] Question: Does the paper provide open access to the data and code, with sufficient instructions to faithfully reproduce the main experimental results, as described in supplemental material?
    \item[] Answer: \answerYes{} % Replace by \answerYes{}, \answerNo{}, or \answerNA{}.
    \item[] Justification: Code will be released upon acceptance.

\item {\bf Experimental setting/details}
    \item[] Question: Does the paper specify all the training and test details (e.g., data splits, hyperparameters, how they were chosen, type of optimizer, etc.) necessary to understand the results?
    \item[] Answer: \answerYes{} % Replace by \answerYes{}, \answerNo{}, or \answerNA{}.
    \item[] Justification: See Sec.\ref{sec:experiments} and Appendix \ref{appx:hyperparameters}.

\item {\bf Experiment statistical significance}
    \item[] Question: Does the paper report error bars suitably and correctly defined or other appropriate information about the statistical significance of the experiments?
    \item[] Answer: \answerNo{} % Replace by \answerYes{}, \answerNo{}, or \answerNA{}.
    \item[] Justification: We follow the common practice by the literature of diffusion model fine-tuning and do not include error bars due to high computation burden.

\item {\bf Experiments compute resources}
    \item[] Question: For each experiment, does the paper provide sufficient information on the computer resources (type of compute workers, memory, time of execution) needed to reproduce the experiments?
    \item[] Answer: \answerYes{} % Replace by \answerYes{}, \answerNo{}, or \answerNA{}.
    \item[] Justification: See Sec.\ref{sec:experiments}.
    
\item {\bf Code of ethics}
    \item[] Question: Does the research conducted in the paper conform, in every respect, with the NeurIPS Code of Ethics \url{https://neurips.cc/public/EthicsGuidelines}?
    \item[] Answer: \answerYes{} % Replace by \answerYes{}, \answerNo{}, or \answerNA{}.
    \item[] Justification: We follow the NeurIPS Code of Ethics.

\item {\bf Broader impacts}
    \item[] Question: Does the paper discuss both potential positive societal impacts and negative societal impacts of the work performed?
    \item[] Answer: \answerYes{} % Replace by \answerYes{}, \answerNo{}, or \answerNA{}.
    \item[] Justification: See Appendix \ref{appx:impact_statement}.
    
\item {\bf Safeguards}
    \item[] Question: Does the paper describe safeguards that have been put in place for responsible release of data or models that have a high risk for misuse (e.g., pretrained language models, image generators, or scraped datasets)?
    \item[] Answer: \answerNo{} % Replace by \answerYes{}, \answerNo{}, or \answerNA{}.
    \item[] Justification: We do not release any new models, and the data we used are publicly accessible.

\item {\bf Licenses for existing assets}
    \item[] Question: Are the creators or original owners of assets (e.g., code, data, models), used in the paper, properly credited and are the license and terms of use explicitly mentioned and properly respected?
    \item[] Answer: \answerYes{} % Replace by \answerYes{}, \answerNo{}, or \answerNA{}.
    \item[] Justification: We cite the original dataset papers.

\item {\bf New assets}
    \item[] Question: Are new assets introduced in the paper well documented and is the documentation provided alongside the assets?
    \item[] Answer: \answerNA{} % Replace by \answerYes{}, \answerNo{}, or \answerNA{}.
    \item[] Justification: This paper does not release any new assets.

\item {\bf Crowdsourcing and research with human subjects}
    \item[] Question: For crowdsourcing experiments and research with human subjects, does the paper include the full text of instructions given to participants and screenshots, if applicable, as well as details about compensation (if any)? 
    \item[] Answer: \answerNA{} % Replace by \answerYes{}, \answerNo{}, or \answerNA{}.
    \item[] Justification: This paper does not involve crowdsourcing nor research with human subjects.

\item {\bf Institutional review board (IRB) approvals or equivalent for research with human subjects}
    \item[] Question: Does the paper describe potential risks incurred by study participants, whether such risks were disclosed to the subjects, and whether Institutional Review Board (IRB) approvals (or an equivalent approval/review based on the requirements of your country or institution) were obtained?
    \item[] Answer: \answerNA{} % Replace by \answerYes{}, \answerNo{}, or \answerNA{}.
    \item[] Justification: This paper does not involve crowdsourcing nor research with human subjects.

\item {\bf Declaration of LLM usage}
    \item[] Question: Does the paper describe the usage of LLMs if it is an important, original, or non-standard component of the core methods in this research? Note that if the LLM is used only for writing, editing, or formatting purposes and does not impact the core methodology, scientific rigorousness, or originality of the research, declaration is not required.
    %this research? 
    \item[] Answer: \answerNA{} % Replace by \answerYes{}, \answerNo{}, or \answerNA{}.
    \item[] Justification: The core method development in this research does not involve LLMs as any important, original, or non-standard components.

\end{enumerate}

\end{document}